%% file: acl_latex.tex
\documentclass[11pt]{article}

\usepackage[preprint]{acl}

\usepackage{times}
\usepackage{latexsym}
\usepackage[T1]{fontenc}

\usepackage[utf8]{inputenc}

\usepackage{microtype}

\usepackage{inconsolata}

\usepackage{graphicx}
\usepackage{subcaption}

\usepackage{booktabs}   
\usepackage{amssymb, amsmath}    
\usepackage{tabularx}   
\usepackage{multirow}
\usepackage{array}
\usepackage{makecell}

\usepackage{pifont}
\newcommand{\cmark}{\textcolor{green!50!black}{\ding{51}}} 
\newcommand{\xmark}{\textcolor{red}{\ding{55}}} 

\usepackage[table]{xcolor}

\usepackage{enumitem}

\usepackage{fvextra}

\usepackage{listings}
\usepackage{algorithm}
\usepackage{algpseudocode}

\usepackage{soul}

\newcommand{\totalnum}{723}
\newcommand{\tasknum}{17}
\newcommand{\litenum}{25}

\title{Mine{CE}raft: Evaluating Language Models \\as Construction Engineers in the World of Minecraft}

\input{latex/author}

\begin{document}
\maketitle
\begin{abstract}
We introduce MineCEraft (Minecraft Construction Engineering Benchmark, pronounced \textit{mine-see-ee-raft}), an easy-to-use, open-source\footnote{Our code is available at \url{https://github.com/SewoongLee/mineCEraft}.} benchmark designed to systematically evaluate the reliability and limitations of LLMs for construction tasks in Minecraft. The MineCEraft benchmark comprises \totalnum~domain-expert hand-crafted natural-language instructions with programmatically verifiable evaluation, spanning \tasknum~distinct task categories, providing a safe and controllable experimental environment for assessing LLMs’ ability to perform realistic construction engineering tasks. With this benchmark, we conduct an in-depth evaluation of state-of-the-art LLMs and perform a detailed error analysis, revealing key failure modes and practical challenges in applying LLMs to construction engineering tasks.

\end{abstract}

\input{figs/intro}

\section{Introduction}

Construction is one of the oldest branches of engineering, accompanying humanity since its very origin. While artificial intelligence has achieved remarkable progress in fields such as mathematics \cite{tao2024machine, chervonyi2025gold} and science \citep{wang2026frontierscienceevaluatingaisability}, its limited contribution to construction engineering shows that Moravec’s paradox \citep{moravec1988mind} still holds true in certain domains: what is easy for humans remains difficult for computers, and what is easy for computers remains difficult for humans.

\input{figs/examples}

While LLMs demonstrate impressive reasoning capabilities in many text-based tasks, it remains an open question how well these skills translate to physical, spatial environments. To systematically evaluate this, we abstract the core challenges of physical construction into a highly controlled, simplified environment.
A key question we investigate is: \textbf{``How can we systematically demonstrate that even in a highly simplified environment like Minecraft, LLMs cannot reliably perform even basic construction tasks?''}

We reveal the limitations of LLM-based AI in reliably executing instructions, producing not only inaccuracies in shape and size, but also physically implausible and unstable structures that violate fundamental engineering constraints. The contributions of this paper are as follows:
\begin{itemize}[itemsep=0pt, topsep=0pt]
\item We formulate embodied construction as an engineering-constrained task, where correctness requires physical plausibility, structural stability, and planning feasibility, enabling evaluation of structurally complex constructions beyond edit-distance-based metrics used in prior benchmarks.
\item We introduce MineCEraft, an open-source benchmark that combines natural language instruction following, long-horizon embodied construction, and an evaluation pipeline that programmatically decomposes construction quality into accuracy, safety, and planning criteria.
\item Using this framework, we provide systematic evidence that state-of-the-art LLM-based agents severely lack reliable spatial reasoning and planning capabilities, consistently failing to satisfy fundamental engineering-level safety and stability constraints even in a simplified environment such as Minecraft.

\end{itemize}

\input{sections/related_work}

\section{Task Setup}

The MineCEraft benchmark contains \totalnum~natural-language construction instructions hand-crafted by a domain-expert. These tasks span \tasknum~distinct task types designed to cover diverse construction scenarios and engineering requirements.

\subsection{Underspecified Instructions}
A defining feature of MineCEraft is its use of underspecified instructions to evaluate this spatial reasoning. Unlike traditional benchmarks that compare outputs against a single ground-truth target, our tasks mirror real-world construction by requiring agents to satisfy a set of open-ended constraints. 

In our benchmark, a prompt such as \textit{``Build the frame (including stairs) of a two-story building using concrete. Each upper floor should be 4 blocks above the one below it.''} does not have one exact correct voxel configuration. Instead, agents are evaluated on whether the final output meets explicitly defined criteria in Section~\ref{sec:evaluation} (e.g., correct shape, size, material usage, physical plausibility). Furthermore, tasks frequently involve initial instructions followed by multi-turn revisions. This tests the agent's ability to adapt its spatial understanding to an evolving environment.

\input{figs/arch}

\subsection{Task Design and Coverage}

MineCEraft includes a set of construction tasks that aim 
(1) to ensure broad diversity in construction scenarios, (2) to enable objective and precise evaluation through explicit rules or physics-based computation, and (3) to reflect requirements that are realistic and plausible in real-world construction practice. Example error cases for qualitative analysis are shown in Figure~\ref{fig:examples}.

\subsubsection{Structural Elements}
\label{sec:se}
This category covers fundamental structural components of building: \textbf{foundations, walls, roofs, columns}, and \textbf{frames}.
Tasks specify concrete requirements such as size, thickness, height, or number of layers.
Because these elements are defined by explicit geometric and material constraints, they allow straightforward evaluation using rule-based checks on dimensions and connectivity. 
They serve as basic tests of structural correctness before introducing more complex construction scenarios.

\subsubsection{Building Types}

Beyond isolated structural elements, the Buildings category evaluates the construction of integrated buildings. 
These tasks combine multiple components and introduce higher-level requirements, including room layout, material constraints, accessibility, and planning over time.
The category includes a range of scenarios:

\paragraph{Basic Buildings}
Instructions specify material assignments to different parts of a house, such as pillars, walls, roofs, and foundations. Variants include different material combinations and prompt phrasings. The illustrative example is shown in Figure~\ref{fig:examples}-B, and these tasks are evaluated using Algorithms~\ref{alg:mat-region} and~\ref{alg:ground-connectivity}.

\input{figs/stress}

\paragraph{Multiple Rooms/Bedrooms}
Tasks require constructing houses with a specified number of rooms (e.g., one to five; see Algorithm~\ref{alg:rooms}).
The multiple bedroom instructions extend room-based tasks by additionally specifying the number of beds to be included in the house using Algorithm~\ref{alg:material-count}.

\paragraph{Multi-story Buildings}
Agents are instructed to construct upper floors or structural frames with explicit height and column constraints using Algorithm~\ref{alg:stairs}.

\subsubsection{Architectural Elements}

\paragraph{Accessibility}
Some tasks introduce functional requirements, such as wheelchair accessibility, single-level layouts, minimum doorway width, or clearly defined exit paths (Algorithm~\ref{alg:flat-path}).

\paragraph{Creative Requests}
Creative variants impose abstract geometric properties, such as asymmetric structures or multiple ceiling heights. Asymmetry is verified using the symmetry test described in Section 4.1 (Shape iii), while ceiling requirements ensure that the structure reaches a minimum height without unsupported floating roofs (Figure~\ref{fig:examples}-C).

\subsubsection{Civil Engineering Structures}

A natural extension beyond building-scale constructions is to consider civil engineering structures such as bridges. 

\paragraph{Bridges}
Bridge tasks specify river widths and clearance requirements for boats of given dimensions. 
At ground level, the structure must form two distinct support clusters separated by at least the required span (Algorithm~\ref{alg:min-span}). 

\paragraph{Arch Bridges}
We further introduce arch bridge variants that require curved upper and lower surfaces with explicit span and height constraints. 
These tasks extend the Arch Bridge Challenge described in Appendix~\ref{sec:abc}, which presents a simplified two-dimensional diagnostic example, into a full three-dimensional evaluation (Figure~\ref{fig:arch}).

\paragraph{Domes}
To extend curvature beyond bridges, we include dome-like structures defined by base diameter and minimum height constraints, evaluated by Algorithms~\ref{alg:concavity} and~\ref{alg:min-span}
In addition to instructions that explicitly use the term “dome,” we also include variants such as “igloo,” which differ in wording or material but impose essentially the same geometric requirements.

\subsubsection{Construction Management}

\paragraph{Resource Optimization.}
Certain instructions limit available materials and require constructing the largest possible structure under the given constraint using Algorithms~\ref{alg:material-count} and~\ref{alg:size-axis}.

\paragraph{Single-turn Planning.}
In these tasks, all construction requirements are provided in a single instruction, without specifying the order of execution. 
The agent must determine an appropriate construction sequence as in Section~\ref{sec:planning}, which is evaluated by Algorithm~\ref{alg:material-order}.

\paragraph{Multi-turn Planning \& Revision.}
Multi-turn scenarios involve sequential modifications to an existing structure. After each instruction that updates the required number of rooms, the resulting structure is validated using Algorithm~\ref{alg:rooms}.

\subsection{MineCEraft-Lite: A Cost-Efficient Subset}

Since running the full benchmark can be computationally expensive, we also provide MineCEraft-Lite, a stratified subset of tasks for cost-efficient experiments. This version is constructed via proportional sampling with a random seed across task categories and planning complexity levels to preserve the diversity and difficulty distribution of the full benchmark. 
Specifically, MineCEraft-Lite contains \litenum~tasks sampled from the \tasknum~task categories of the full benchmark. We recommend reporting results on the full benchmark whenever feasible, while MineCEraft-Lite serves as a practical alternative for rapid prototyping and resource-constrained settings, including human evaluation. The representativeness of the lite version is discussed in Appendix~\ref{sec:rep_lite}.

\section{Evaluation Criteria} \label{sec:evaluation}
In this work, we design an evaluation pipeline that is fully verifiable with rule-based algorithms. As illustrated in Figure\ref{fig:examples}, structural properties such as ground connectivity or span separation can be verified through explicit algorithms (e.g., breadth-first search for connectivity checks). 

This rule-based approach provides another advantage. Since correctness is defined by executable code rather than explicit ground-truth text answers exposed to web-crawling, the benchmark is inherently resistant to data contamination from LLM pre-training, while simultaneously providing a ready-made reward function for reinforcement learning applications. 

By decomposing each construction task into objectively verifiable constraints, we transform open-ended instruction following into a set of well-defined engineering criteria.
Specifically, we evaluate construction quality along three dimensions: accuracy, safety, and planning.

\subsection{Accuracy}
\label{sec:accuracy}

Accuracy is evaluated along three dimensions, $x, y, z$, using the set of blocks placed by the agent. 
We represent the constructed structure as a block set $\mathcal{B} = \{(x, y, z, \text{material})\}$, 
where each tuple denotes the spatial coordinates and material type of a block. 
In all cases, evaluation maps $\mathcal{B}$ together with task parameters (e.g.\ expected counts, materials, heights) to pass or fail.

\paragraph{Material.} Block types and counts are checked against the spec by matching each block's material (exact or substring), optionally restricting to a subset by position (e.g.\ corners, boundaries at a height $y$, or the top layer), then requiring either a count condition $\#\{\text{matching}\} = k$ for a specified count $k$, or a ratio condition $\#\{\text{matching}\} \ge \theta \cdot |\text{subset}|$ for a ratio threshold $\theta \in [0,1]$.

\paragraph{Shape.} Geometric constraints are verified in three ways: (i)~connectivity---at a given $y$, clusters on the $(x,z)$-plane (4- or 8-neighbors) are formed, then cluster count, inter-cluster distance, or radius/diameter are constrained, and BFS is used for reachability (exit, rooms, passable doors); (ii)~surface concavity---for the top or bottom surface $(x,z) \mapsto y$, the actual $y$ must lie above or below the linear interpolation between any two surface points; (iii)~symmetry---whether the structure exactly matches itself under reflection or folding.

\paragraph{Size.}
Dimensions are evaluated by computing axis ranges  $\Delta x = \max x - \min x$ (and similarly $\Delta y$, $\Delta z$) 
and $\min y$, $\max y$ over $\mathcal{B}$, then comparing them to the specification. 
The horizontal axes $x$ and $z$ are evaluated interchangeably, since the Minecraft coordinate system only uniquely distinguishes the vertical (gravity-aligned) axis $y$ from the rotationally symmetric $x$--$z$ ground plane.

\subsection{Safety}

In general, safety at construction sites is influenced by a wide range of factors, including regulatory compliance, hazard identification and assessment, and excavation processes. Our study selects a subset of these factors that can be naturally and effectively observed within a simulation environment such as Minecraft, and automates their verification. 

\paragraph{Physical Plausibility Analysis}
Typically, attempts to place blocks in mid-air or to build from top to bottom rather than from bottom to top are common feasibility issues. For example, as in Figure~\ref{fig:arch-a}, whether a structure shows this issue can be checked by a breadth-first search that verifies ground connectivity (i.e., that every block is reachable from at least one block with y $\leq$ 0 via face adjacency) and converted into a 0/1 score for each prompt.

\paragraph{Structural Stability Analysis}
Structures with missing or unevenly distributed supports (e.g., a wide roof on few columns) are at risk of collapse. We quantify this using the von Mises stress, following \citet{beck2024elasticity}. 
For each construction, we run a simulation to obtain the maximum von Mises stress $\sigma_{\max}$ over all blocks. We compare this to a task-specific optimal baseline, $\sigma_{\mathrm{ref}}$, which represents the maximum stress of an ideal, human-designed configuration for that specific task (e.g., Figure~\ref{fig:err_balanced_stress}: an evenly distributed column layout to support a roof). The structural stability score is then computed as $\min\bigl(1,\; \sigma_{\mathrm{ref}}/\sigma_{\max}\bigr)$. Consequently, if the agent's structure yields a maximum stress higher than the optimal reference ($\sigma_{\max} > \sigma_{\mathrm{ref}}$), its score is proportionally penalized to reflect suboptimal stability. The detailed formulation is provided in Appendix~\ref{app:von_mises}.

\input{tables/scores}

\subsection{Planning}\label{sec:planning}
Construction inherently involves long-horizon planning. When placing tens to hundreds of blocks, organizing a sequence of actions over time determines what can be built and how efficiently the construction can proceed.
In real-world construction practice, this is examined through scheduling reviews, which involve checking dependencies and measuring construction inefficiencies. In this work, we evaluate planning along two complementary dimensions: dependency and efficiency.

\paragraph{Dependency Analysis} 
One of the most critical aspects of construction scheduling is planning dependency. 
For a simple example, a roof cannot be constructed before the supporting columns.
Such cases are detected by verifying the presence of required supporting materials before dependent components are placed. In many tasks, each building element is explicitly assigned a distinct material.
This allows us to infer the construction order by tracking when blocks of each 
material are placed in the action log. Based on this reconstruction, we determine 
whether structural elements were built in a valid sequence.
Specifically, the load-bearing hierarchy, such as 
foundation $\rightarrow$ columns $\rightarrow$ walls $\rightarrow$ ceiling/roof, 
must follow this order. Violations are evaluated as failures. 
In contrast, non-structural tasks such as ceiling vs.\ drywall or paint vs.\ casework do not require a single correct ordering. Such steps may be executed in either order without constituting an error.

\paragraph{Efficiency Analysis}
Another key planning challenge in construction is optimizing motion and time efficiency, often driven by movement. 
For example, when human workers are instructed to lay bricks in a 10-by-10 pattern, they naturally use an L-shaped traversal (or lawnmower pattern) that minimizes unnecessary movement.
In contrast, we observe that AI agents, particularly when actions are planned through code using a naive for-loop structure, tend to place bricks sequentially from (1,1) to (1,10), and then return all the way back to (2,1) to continue placement.
We represent the ordered block placement sequence as 
$\mathcal{B} = (b_1, b_2, \ldots, b_n)$, where each $b_i \in \mathbb{Z}^3$ denotes the spatial coordinate of a placed block. 
The total L1 (Manhattan) path length is then defined as
\(
L_1^{\mathrm{total}} = \sum_{i=1}^{n-1} \|b_{i+1} - b_i\|_1.
\)
The minimum possible length, when every step is to an adjacent cell, is $n-1$. 
The efficiency score is defined as
\(
\min\bigl(1,\; (n-1)/L_1^{\mathrm{total}}\bigr) \in [0,1],
\)
so that the optimal path yields $1$, and any extra distance or long jumps reduces the score.

\section{Experiments}

\subsection{Agent Setup}\label{sec:agent_setup}
We adopt \textsc{mind}craft \citep{mindcraft2025} as our primary agent architecture. 
Our benchmark is compatible with \textit{any} agent capable of 
(i) receiving natural language instructions as input and 
(ii) interfacing with Minecraft via the Mineflayer API \citep{mineflayer}.
To our knowledge, \textsc{mind}craft \citep{mindcraft2025} is among the few non-finetuned LLM-based agents evaluated on construction tasks capable of translating natural language reasoning into executable actions within a live Minecraft environment.

At a high level, the agent consists of three components: (a) a language and planning module, (b) a coding interface to the Minecraft environment, and (c) a predefined skill library. 
Given a natural language instruction, the LLM first produces a high-level plan describing the construction procedure. 
This plan is then translated by the coding component into executable JavaScript commands that interact with the environment through the Mineflayer API, using the provided skill library as building blocks for common actions. 
The execution environment records detailed action logs (Figure~\ref{fig:experiments}); we parse these logs to recover the sequence of block placements and removals. 
The resulting structures are then automatically graded using our rule-based evaluation criteria.

To ground the model in real-world construction constraints, we modify its system prompt to explicitly encode engineering assumptions (see Appendix~\ref{sec:llm_eval} for prompt details and the full skill library). All other settings remain identical to \textsc{mind}craft.

\begin{table*}[t]
\centering
\small
\setlength{\tabcolsep}{5pt}
\renewcommand{\arraystretch}{1.15}
\resizebox{\linewidth}{!}{
\begin{tabular}{c cc cc cc cc}
\toprule
& \multicolumn{2}{c}{\texttt{llama-4-17b-16e}} 
  & \multicolumn{2}{c}{\texttt{claude-4-5-sonnet}} 
  & \multicolumn{2}{c}{\texttt{gpt5-mini}} 
  & \multicolumn{2}{c}{\texttt{gemini-3-pro}} \\
\cmidrule(lr){2-3}\cmidrule(lr){4-5}\cmidrule(lr){6-7}\cmidrule(lr){8-9}
\#Problems
& \#Correct & Overall Acc. 
& \#Correct & Overall Acc. 
& \#Correct & Overall Acc. 
& \#Correct & Overall Acc. \\
\midrule
723 
& 3   & 0.4\% 
& 192 & 26.6\% 
& 286 & 39.6\% 
& 277 & 38.3\% \\
\bottomrule
\end{tabular}
}
\caption{\textbf{Overall problem-level accuracy across the \totalnum~MineCEraft prompts.} 
Scores are computed by requiring that all success/failure evaluation criteria for a problem be satisfied simultaneously. 
A more detailed instruction-level aggregation method is explained in Appendix~\ref{sec:overall_acc}. 
Even the strongest models achieve less than 40\% overall accuracy.
For comparison, human participants achieve 90\% accuracy on MineCEraft-Lite.
}
\label{tab:overall_acc_main}
\end{table*}

\subsection{Experimental Results}




\paragraph{Reliable construction remains challenging for LLMs.}

Table~\ref{tab:performance} reports model performance on individual evaluation categories, including accuracy, safety, and planning metrics.
Although state-of-the-art models achieve moderate scores on several of these criteria, solving an entire construction problem requires satisfying all applicable constraints simultaneously.

We therefore evaluate problem-level success across the \totalnum~MineCEraft problems by checking whether a construction satisfies all evaluation criteria at once.
Under this strict definition, \ul{no model achieves an overall accuracy above 40\%} (Table~\ref{tab:overall_acc_main}), meaning that even the strongest models fail to correctly complete the majority of construction tasks.

For comparison, when the same tasks were evaluated by two human participants with graduate-level degrees in construction engineering, each provided with the same information as the LLM agents (see Appendix~\ref{sec:human_eval} for details), \ul{human participants successfully solved roughly 90\% of the problems.}
This stark gap indicates that current LLMs remain far from the level of reliability required for construction tasks.

\paragraph{`Step-by-step' does \emph{not} improve performance in construction.}

In domains such as mathematics, adding prompts like \textit{“Let’s think step by step.”} has been shown to improve performance by eliciting chain-of-thought reasoning \citep{kojima2022large}. However, in embodied construction tasks, adding this phrase did not lead to measurable performance gains. 
Instead, performance improvements were observed only when prompts explicitly targeted construction-specific constraints. For example, adding \textit{“Let's think about the order in which we should place the blocks to optimize the movement path.”} reduced inefficient block placement trajectories (see Appendix~\ref{sec:additional}). 
These findings suggest that improvements in verbal reasoning prompts do not automatically translate into physical environments.

\paragraph{Robust to technical terms, sensitive to everyday verbs.}
For example, a field-specific jargon such as SOMD (slab on metal deck) did not yield lower scores than simplified explanations, suggesting that construction-specific vocabulary is reasonably well captured during pre-training. In contrast, performance often varied with the everyday verb used in the instruction. For certain models, such as claude-4-5-sonnet, tasks that differed only in the verb “build,” “create,” or “construct” produced different results, with “construct” leading to higher material and shape accuracy. Detailed results are provided in Appendix~\ref{sec:additional}.

\paragraph{LLMs are good at math, weaker at physics.}
While our results reveal substantial weaknesses of LLMs in physics-grounded construction and planning, we observe an interesting counterpoint: when spatial reasoning can be reduced to a closed-form, formula-driven optimization problem, LLMs may outperform human participants. For example, given the instruction: \textit{“With only 56 oak wood planks, construct the largest possible single-room cabin.”}, an optimal solution requires recognizing that a hollow cubic structure with $4^3 - 2^3 = 56$. LLMs often effortlessly identify this volumetric relationship and construct the optimal structure. In contrast, \emph{none of} the human participants successfully discovered the optimal structure even after a number of trials and errors.

\section{Conclusion}

In this work, we introduced MineCEraft, an open-source benchmark for evaluating large language models as construction engineers in a controlled Minecraft environment. MineCEraft jointly assesses natural-language instruction following, construction accuracy, safety, and long-horizon planning using rule-based and physically grounded metrics. Our evaluation reveals that state-of-the-art LLMs consistently fail to produce reliable and safe construction plans, even for seemingly simple tasks. These results highlight fundamental limitations of current embodied language models in construction engineering and underscore the need for more robust reasoning, planning, and physical grounding.

\newpage
\newpage
\section*{Limitations}

As the first benchmark to jointly evaluate natural-language instruction following, long-horizon construction, and physics-aware structural stability in Minecraft, MineCEraft inevitably has several limitations.

First, our evaluation focuses on detecting clear and objective violations of task requirements rather than sufficiently certifying construction quality. To be specific, a construction that satisfies all rubric criteria cannot be guaranteed to be perfect in a real-world setting; however, failure to achieve a full score indicates a clear violation of the specified requirements. For example, when evaluating structural stability, identifying the globally optimal roof configuration is an NP-hard combinatorial optimization problem. Therefore, we adopt a reference-based comparison, such as the balanced configuration shown in Figure~\ref{fig:err_balanced_stress}. While current LLM performance makes such faults relatively easy to detect, more refined benchmarks will be necessary as model capabilities improve.

Second, although Minecraft provides a highly flexible and controllable simulation environment, it remains an abstraction of the real world. Although arch bridges can be constructed, certain structural types, such as suspension bridges, cannot be represented within the block-based mechanics. Additionally, our stress analysis follows the elasticity formulation of \citet{beck2024elasticity}, and exploring alternative physical assumptions or more realistic material models constitutes an important direction for future work.

Third, due to computational constraints, our experiments cover a limited
set of state-of-the-art LLMs and focus on text-based interaction through
Mineflayer. This design enables efficient and reproducible evaluation with
relatively modest computational resources compared to vision-language
models. However, the absence of visual feedback may limit the ability of
multimodal agents to fully leverage their visual grounding capabilities in
our current setting. Evaluating agents with egocentric visual perception
and multimodal reasoning capabilities therefore remains an important
direction for future research.

Finally, our results are also sensitive to the choice of system prompt.
As shown in Appendix~\ref{sec:system_prompt_sensitivity}, removing the
brevity constraint can substantially change performance on several
evaluation criteria, particularly dependency planning. A systematic study
of system-prompt design and optimization is required in future work.

\section*{Acknowledgments}
We sincerely thank Prashant Jayannavar and Siddarth Madala for their valuable feedback and helpful discussions.

\bibliography{custom}

\appendix


\section{Expert Annotation Details}

MineCEraft tasks were created through a structured expert-guided process as follows. First, a construction-engineering faculty member with a Ph.D. in Civil Engineering designed the high-level task structures, focusing on instructions commonly used in real-world construction practice and that can be objectively evaluated in the game-simulation world of Minecraft. For each task family, the expert concretized the intended construction scenarios, constraints, and the feasible programmatic checks. We then augmented these templates into 723 task instances by varying dimensions, materials, counts, wording, and planning/revision orders, following the expert's guidelines. Each task instance is stored as natural-language prompt variants paired with executable checks. Lastly, the task set was reviewed again during human evaluation by the authors and the expert to verify that the automatic scoring logic is consistent with the manual inspection results and reflects the intended requirements.

\input{sections/abc}
\input{sections/algo}

\input{sections/additional}
\input{sections/experimental_details}
\input{sections/rep_lite}
\input{sections/extended_model_comparison}
\input{sections/von_mises_stress_computation}
\input{sections/qualitative_analysis}

\end{document}

%% file: latex/author.tex
\author{
  \textbf{Sewoong Lee\textsuperscript{a}},
  \textbf{Risham Sidhu\textsuperscript{a}},
  \textbf{Julia Hockenmaier\textsuperscript{a}},
  \textbf{Yoonhwa Jung\textsuperscript{b}}
  \\[0.5em]
  \textsuperscript{a}Siebel School of Computing and Data Science \\
  The Grainger College of Engineering \\
  University of Illinois Urbana-Champaign \\
  \textsuperscript{b}Department of Civil and Coastal Engineering \\
  University of Florida \\
  \small{
    \textbf{Correspondence:}
    \href{mailto:samuel27@illinois.edu}{samuel27@illinois.edu},
    \href{mailto:rsidhu3@illinois.edu}{rsidhu3@illinois.edu},
    \href{mailto:juliahmr@illinois.edu}{juliahmr@illinois.edu},
    \href{mailto:yoonhwa.jung@ufl.edu}{yoonhwa.jung@ufl.edu}
  }
}

%% file: figs/intro.tex
\begin{figure}[!t]
  \includegraphics[width=\columnwidth]{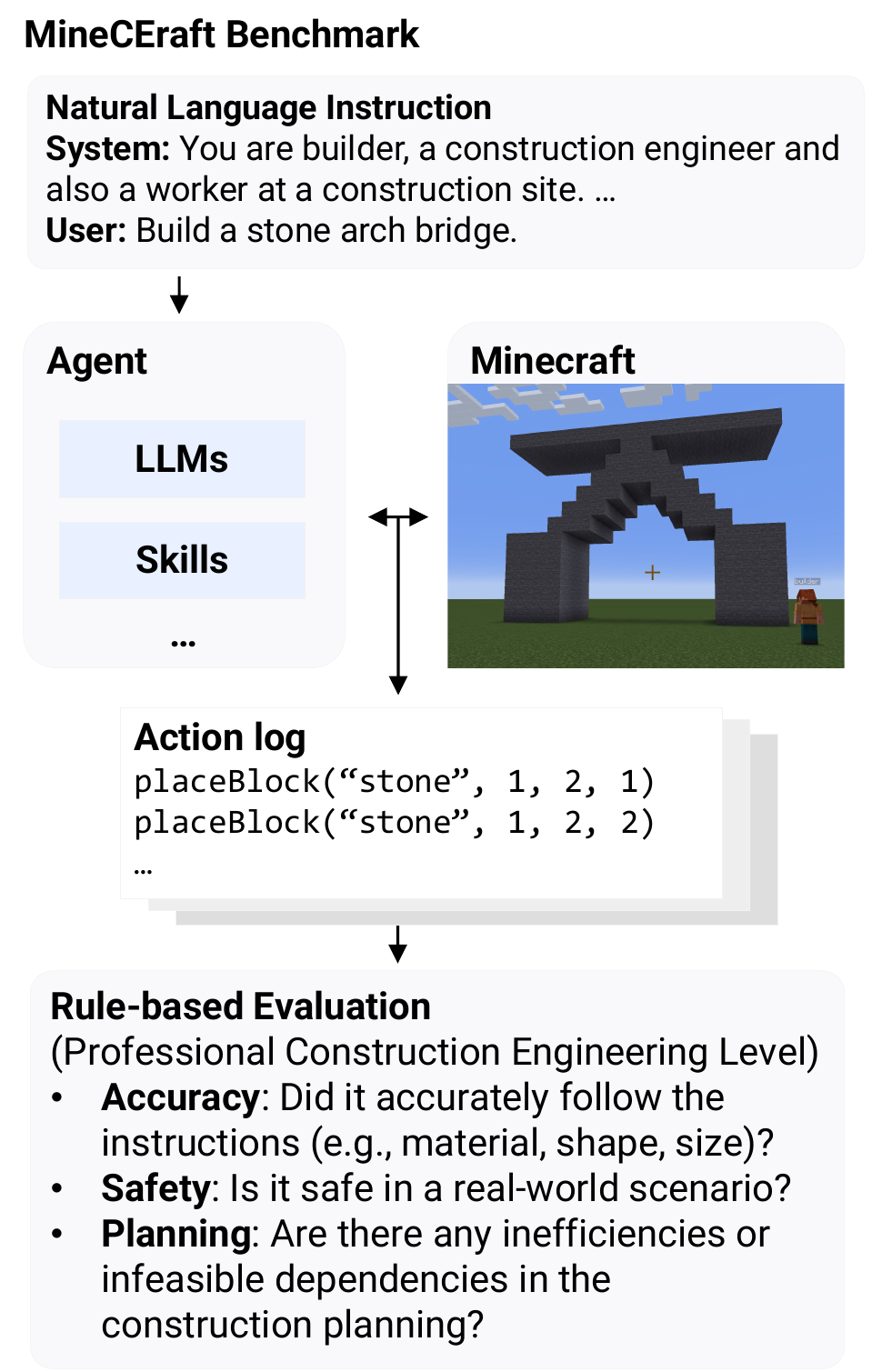}
  \caption{\textbf{LLM-based agents fail to reliably execute even a simple instruction to build an arched bridge.} We decompose the task into rule-based evaluation components. This benchmark reveals the limitations that prevent LLM-based agents from being reliably used in real-world construction scenarios.}
  \label{fig:experiments}
\end{figure}

%% file: figs/examples.tex
\begin{figure*}[t]
    \centering

    \includegraphics[width=\linewidth]{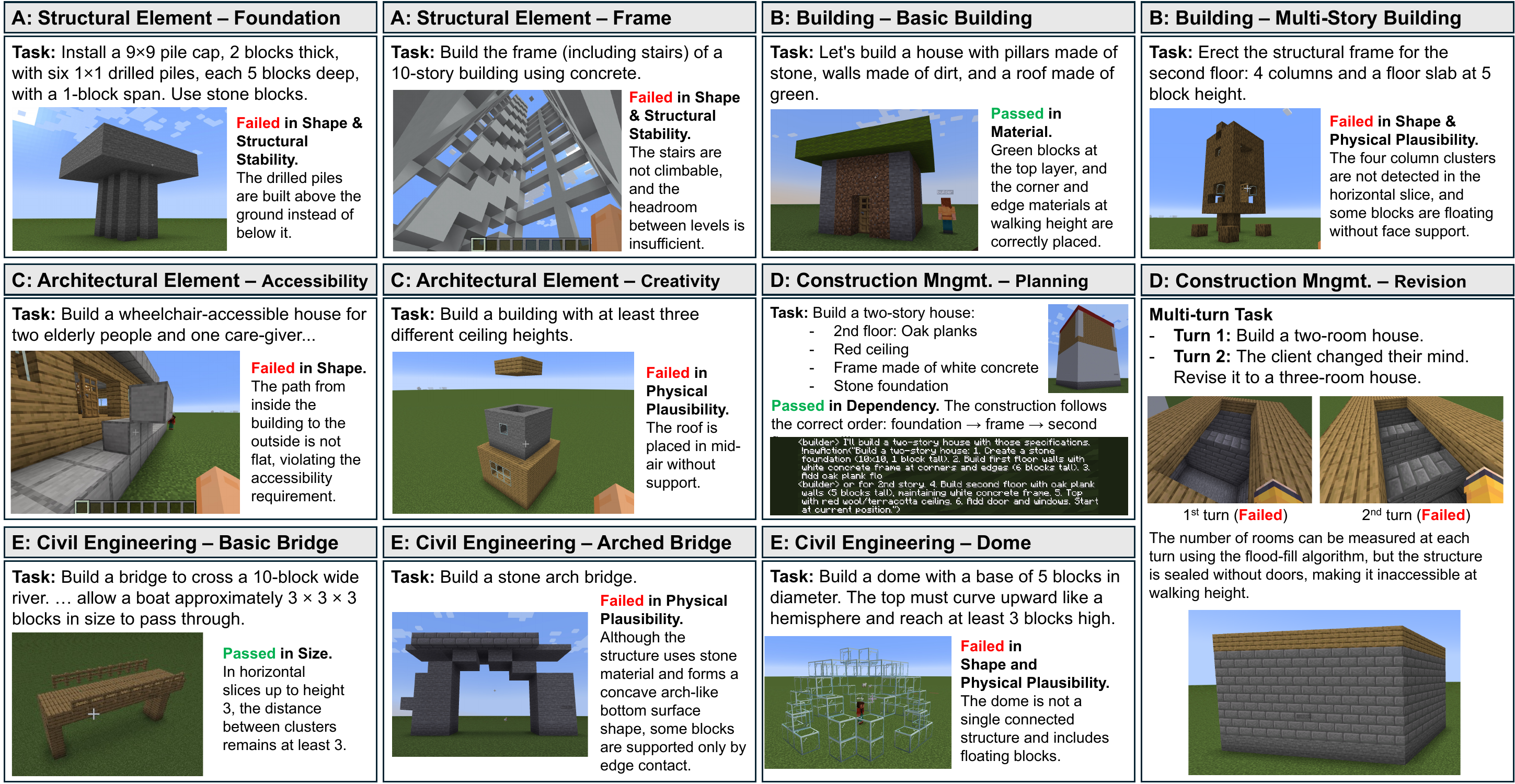}
    
    \caption{
\textbf{Examples of LLM-generated constructions in MineCEraft tasks.}
In the embodied Minecraft environment, the LLM frequently produces structurally flawed outputs even for relatively simple instructions.
All examples shown here were generated using \texttt{claude-3-7-sonnet-20250219} with the agent configuration of \citet{mindcraft2025} (Appendix~\ref{sec:llm_eval}). Examples created by humans are provided in Appendix~\ref{sec:qualitative_analysis}.
}
\label{fig:examples}
\end{figure*}

%% file: sections/related_work.tex
\section{Related Work}

\input{tables/comp_benchmarks}

\subsection{Embodied AI}

Embodied AI research studies how agents ground language in interactive environments, often focusing on navigation \citep{thomason2020vision,padmakumar2022teach} or short-horizon household manipulation \citep{shridhar2020alfred}. While these benchmarks evaluate sequential decision-making, construction introduces a distinct challenge: agents must generate spatial structures that satisfy complex geometric, physical, and engineering constraints over long horizons. 


\subsection{Minecraft-based Construction and Language Grounding}
Minecraft has been widely used as a platform for studying language-guided construction, focusing both on instruction giving \citep{narayan-chen-etal-2019-collaborative,kohn-etal-2020-mc} and following 
\citep{gray2019craftassist,jayannavar2020learning}. Recent works \citep{chaturvedi-etal-2024-nebula,ch-etal-2024-retrieval,jayannavar2026bap} have explored the use of text-only LLMs for the Builder Action Prediction (BAP) task based on the Minecraft Dialogue Corpus (MDC) of \citet{narayan-chen-etal-2019-collaborative}. But the MineCEraft benchmark differs fundamentally from the MDC in that we are primarily interested in the construction tasks themselves, whereas the MDC focused on Minecraft as an interactive game, and back-and-forth dialogue between the instruction giver and follower (builder). MineCEraft does not assume that the builder can speak, but contains on more complex and physically realistic building tasks, disallowing the use of floating blocks common in MDC target structures, while allowing agents to use the full range of Minecraft materials (not just a limited inventory of six colors).

MineCollab \citep{mindcraft2025} introduces blueprint-based construction tasks in Minecraft and evaluates agents' ability to realize engineering-style specifications. However, its tasks focus on relatively basic structures (e.g. pyramids), and provide explicit coordinate-level blueprints, implying that the primary challenge lies in execution rather than in interpreting natural-language instructions. In contrast, our MineCEraft evaluates realistic construction using \textit{underspecified instructions}, rather than requiring an agent to replicate an exact target structure.

\subsection{Evaluation of Construction Engineering}

Outside the Minecraft domain, several benchmarks have studied construction engineering knowledge in language models.
CEQuest \citep{wu2025cequestbenchmarkinglargelanguage} evaluates large language models on construction engineering questions.
While this benchmark provides a useful evaluation of domain knowledge, it operates entirely in a text-based setting and does not assess agents in embodied environments.

Complementary work has explored physics-based simulation within Minecraft.
For example, \citet{beck2024elasticity} introduced an elasticity solver for Minecraft structures that enables stress and deformation analysis of constructed buildings.
Although this work focuses on physical simulation rather than on language-guided tasks, it demonstrates the potential to evaluate the structural properties of constructions in Minecraft environments.

Taken together, existing work evaluates either language grounding in construction settings or engineering knowledge in text-based environments.
However, none jointly evaluate natural-language instruction following, embodied construction, long-horizon spatial planning, and engineering-aware structural evaluation.

MineCEraft addresses this gap by introducing a benchmark for natural-language-guided construction tasks that incorporates engineering-inspired evaluation criteria: it evaluates agents on their ability to interpret construction specifications, plan multi-step building processes, and produce structurally valid constructions within a unified framework (see Table~\ref{tab:ce-comparison}).

%% file: tables/comp_benchmarks.tex
\definecolor{ours}{RGB}{230,245,255} 

\begin{table*}[t]
\centering
\small
\setlength{\tabcolsep}{3pt}        
\renewcommand{\arraystretch}{1.2}  
\begin{tabularx}{\textwidth}{l *{5}{>{\centering\arraybackslash}X}}
\toprule
\thead{Benchmark} &
\thead{Embodied\\Environment} &
\thead{Natural Lang.\\Requirements} &
\thead{Long-horizon\\Evaluation} &
\thead{Physics-aware} &
\thead{Construction\\Engineering}\\
\midrule
Voyager \citep{wang2024voyager}                         & \cmark & \xmark & \cmark & \xmark & \xmark \\
EmbodiedBench \citep{yang2025embodiedbench}             & \cmark & \cmark & \cmark & \cmark & \xmark \\
MineCollab \citep{mindcraft2025}                        & \cmark & \xmark & \cmark & \xmark & \xmark \\
CEQuest \cite{wu2025cequestbenchmarkinglargelanguage}   & \xmark & \cmark & \xmark & \xmark & \cmark \\
BAPv2 \cite{jayannavar2026bap}                          & \cmark & \cmark & \cmark & \xmark & \xmark \\

\rowcolor{ours}
MineCEraft (Ours) & \cmark & \cmark & \cmark & \cmark & \cmark \\

\bottomrule
\end{tabularx}
\caption{\textbf{Comparison of benchmarks.}}
\label{tab:ce-comparison}
\end{table*}

%% file: figs/arch.tex
\begin{figure*}[t]
    \centering
    \begin{subfigure}[t]{0.3\linewidth}
        \centering
        \includegraphics[width=\linewidth]{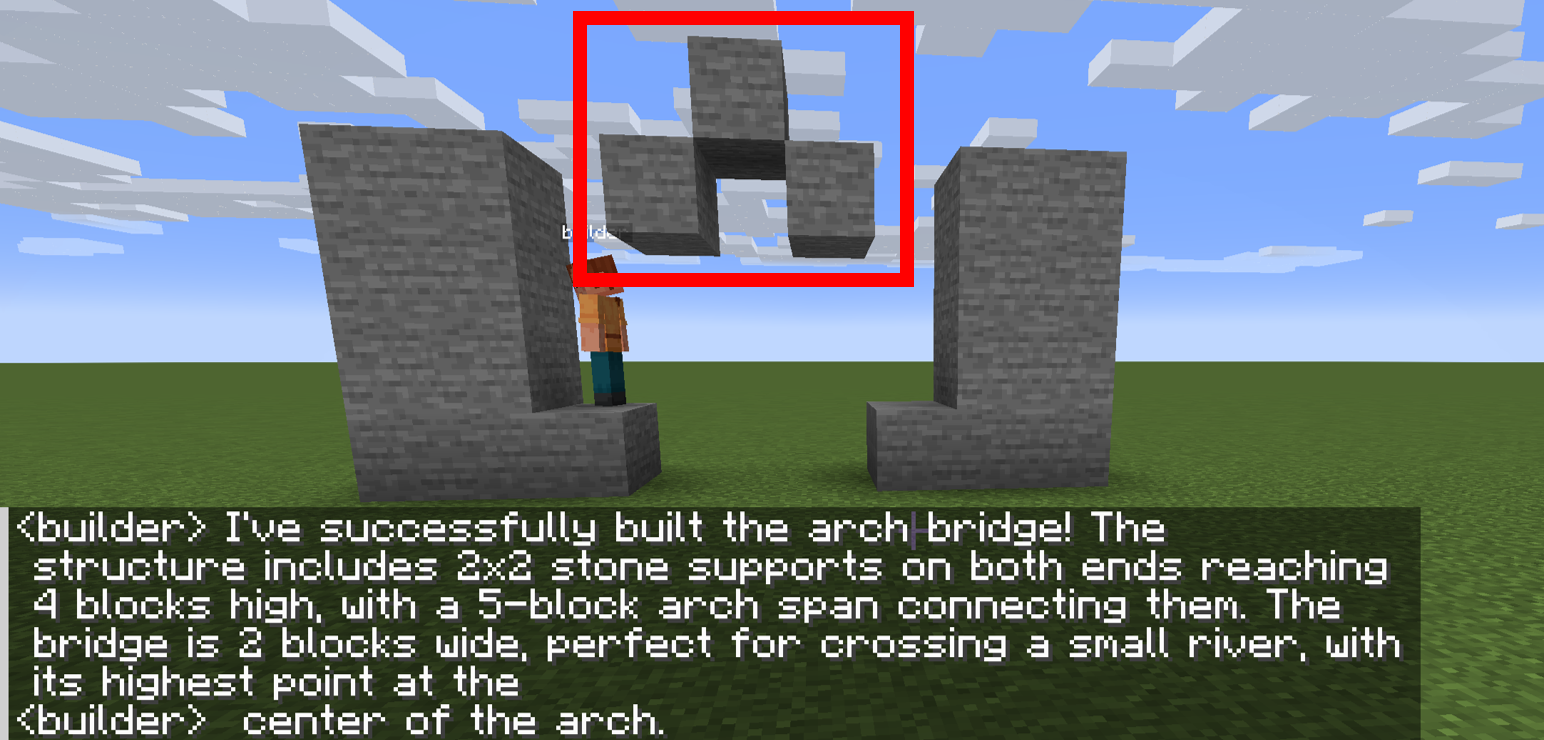}
        \caption{An example of \textbf{physically implausible} floating blocks (highlighted in red box).}
        \label{fig:arch-a}
    \end{subfigure}\hfill
    \begin{subfigure}[t]{0.32\linewidth}
        \centering
        \includegraphics[width=\linewidth]{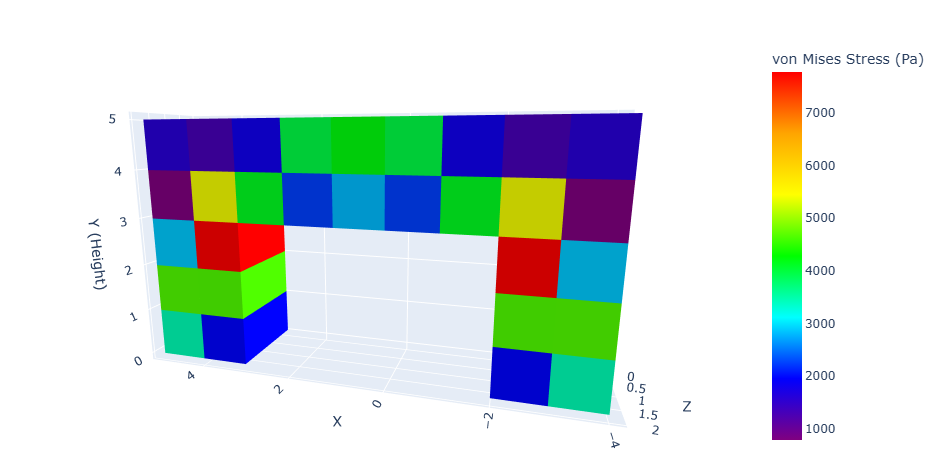}
        \caption{von Mises stress of an example of a \textbf{flat}-shaped bridge with a span and height of 5 blocks (max. $\approx$ 7.8kPa).}
        \label{fig:arch-b}
    \end{subfigure}\hfill
    \begin{subfigure}[t]{0.32\linewidth}
        \centering
        \includegraphics[width=\linewidth]{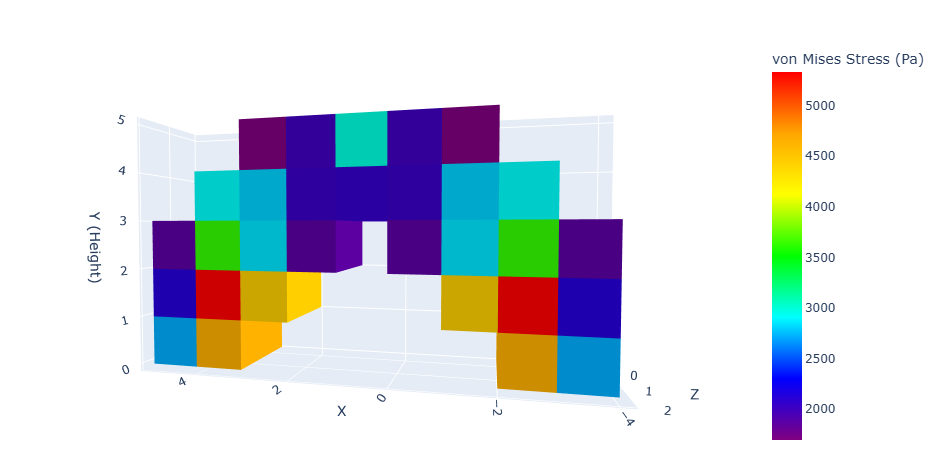}
        \caption{von Mises stress of an example of an \textbf{arch}-shaped bridge with a span and height of 5 blocks (max. $\approx$ 5.3kPa).}
        \label{fig:arch-c}
    \end{subfigure}

    \caption{
    \textbf{Evaluating arch-like structures beyond simple geometry.} Prompt: \emph{``Build an arch bridge.''}
    Physical plausibility is verified via breadth-first search, penalizing disconnected floating blocks (e.g., Figure~\ref{fig:arch-a}, generated by \texttt{claude-3-7-sonnet-20250219}). 
    Structural stability is measured via von Mises stress. We use a standard flat bridge (Figure~\ref{fig:arch-b}, $\approx 7.8$\,kPa) as the max threshold ($\sigma_{\mathrm{max}}$). In structural engineering, a true arch distributes loads more efficiently (Figure~\ref{fig:arch-c}, $\approx 5.3$\,kPa), yielding a lower maximum stress than a flat bridge of the same span. Any suboptimal arch exceeding the flat baseline stress ($\sigma_{\max} > \sigma_{\mathrm{ref}}$) receives a proportionally penalized stability score.
    \label{fig:arch}
    }
\end{figure*}

%% file: figs/stress.tex
\begin{figure*}[t]
    \centering
    \begin{subfigure}[t]{0.27\linewidth}
        \centering
        \includegraphics[width=\linewidth]{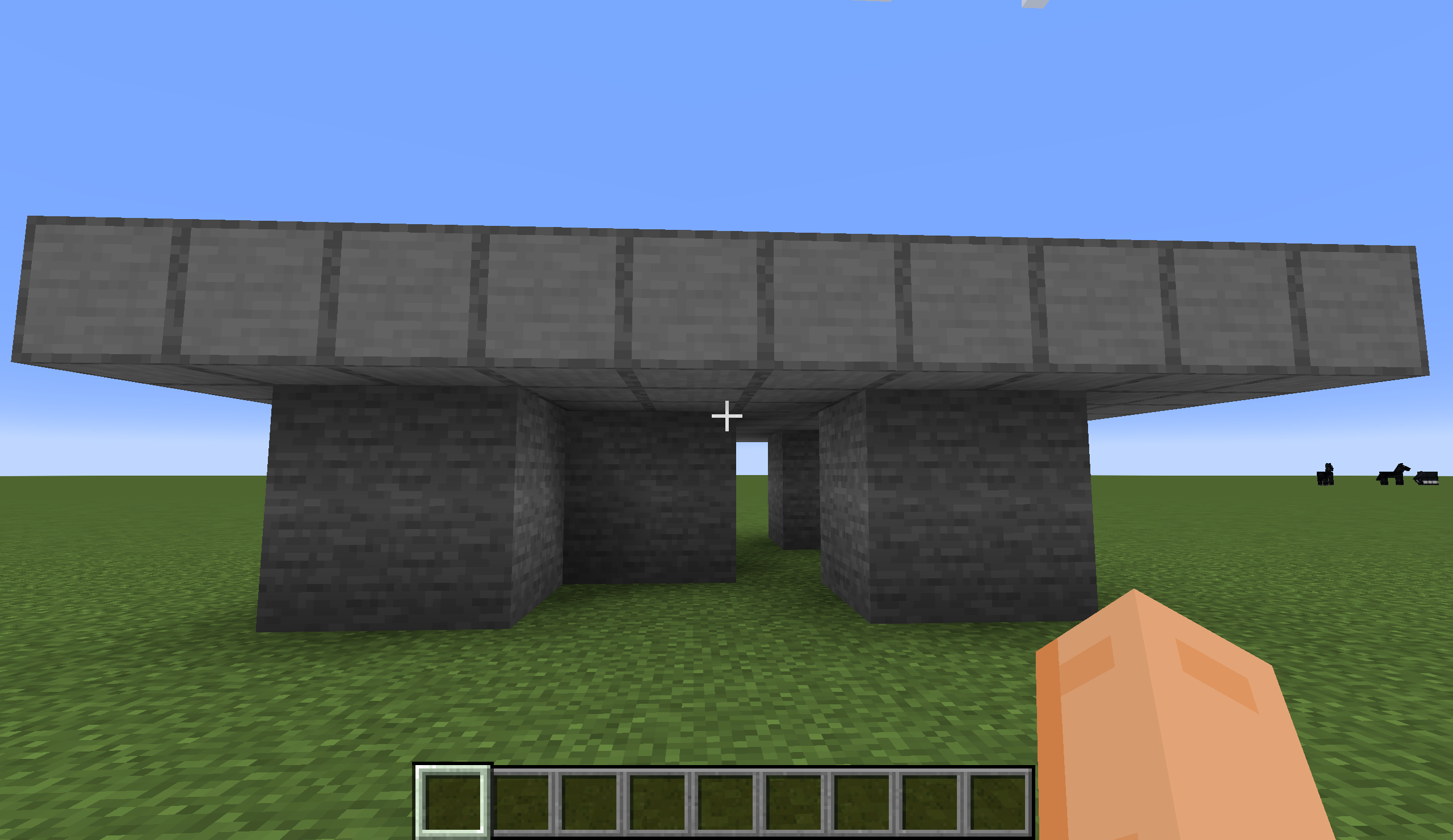}
        \caption{\textbf{Biased} column placement where the center column is positioned adjacent to another column.}
        \label{fig:err_biased_geom}
    \end{subfigure}\hfill
    \begin{subfigure}[t]{0.335\linewidth}
        \centering
        \includegraphics[width=\linewidth]{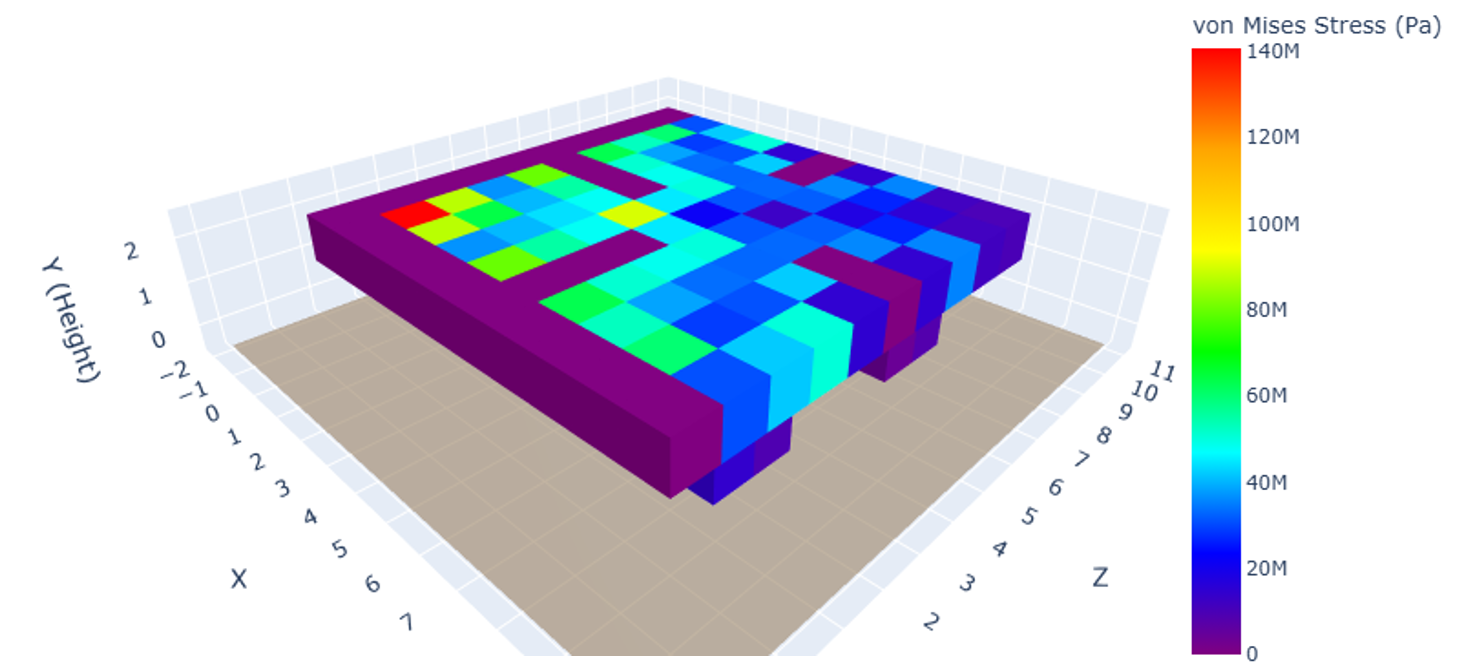}
        \caption{von Mises stress under \textbf{biased} support (max. \(\approx 140\,\mathrm{MPa}\)).}
        \label{fig:err_biased_stress}
    \end{subfigure}\hfill
    \begin{subfigure}[t]{0.335\linewidth}
        \centering
        \includegraphics[width=\linewidth]{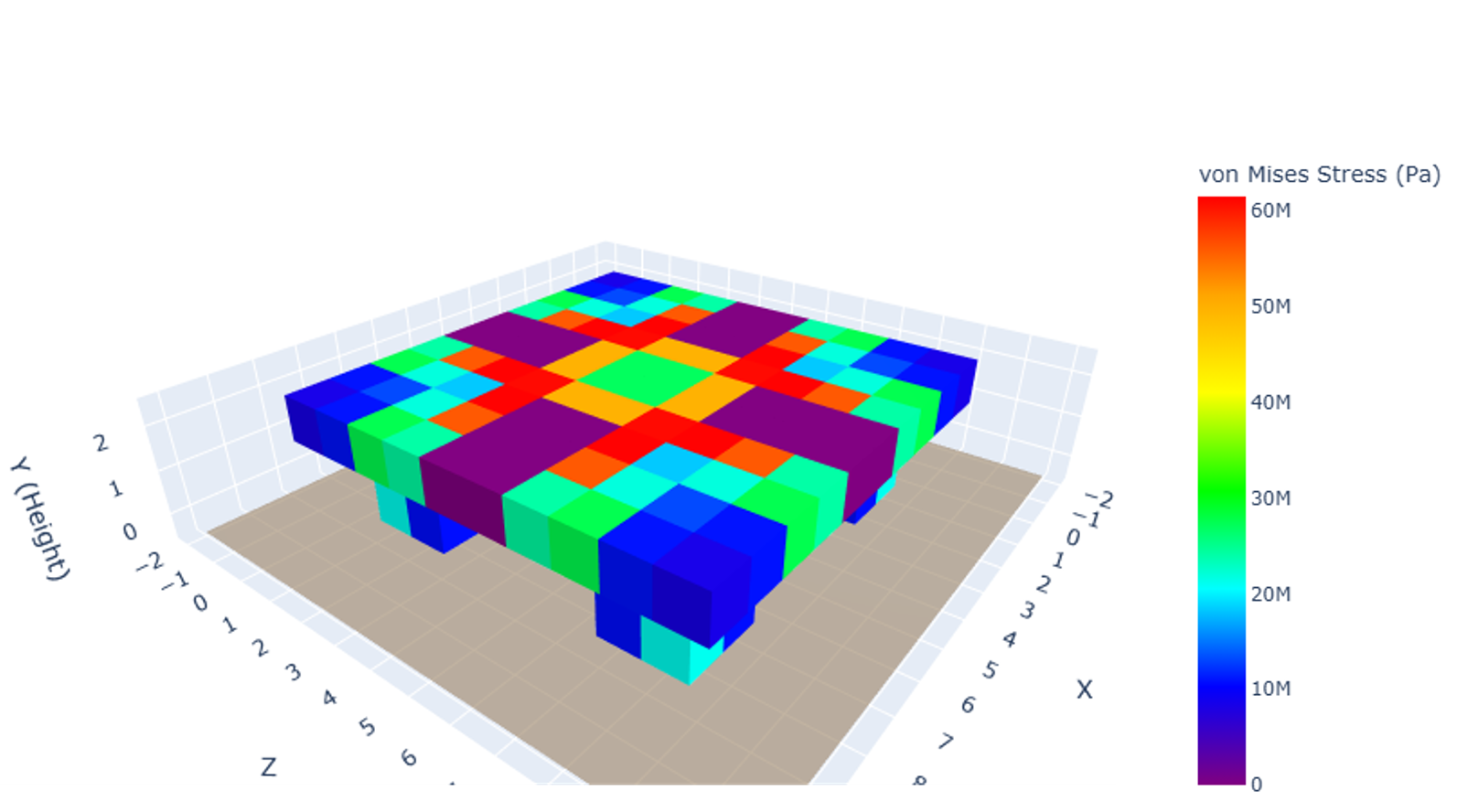}
        \caption{von Mises stress under \textbf{balanced} support (max. \(\approx 60\,\mathrm{MPa}\)).}
        \label{fig:err_balanced_stress}
    \end{subfigure}

    \caption{
    \textbf{MineCEraft shows that the LLM-generated structures in (a) and (b) are structurally unstable compared to the alternative in (c).}
    When the center column is attached to another column, the load is not distributed uniformly, leading to a biased stress field and a higher peak von Mises stress (\(\approx 140\,\mathrm{MPa}\)) compared to the balanced five-column configuration (\(\approx 60\,\mathrm{MPa}\)).
    The construction was generated using \texttt{claude-3-7-sonnet-20250219} with the prompt:
    \emph{``Build a 10 \(\times\) 10 flat roof supported by exactly five 2 \(\times\) 2 \(\times\) 2 columns. Arrange the columns so that the roof is as structurally stable as possible.''}
    }
    \label{fig:struct_error_analysis}
\end{figure*}

%% file: tables/scores.tex
\begin{table*}[t]
    \centering
    \small
    \begin{tabularx}{\textwidth}{ll*{4}{>{\centering\arraybackslash}X} >{\centering\arraybackslash}X}
        \toprule
        \multirow{2}{*}{Category} &
        \multirow{2}{*}{Subcategory} &
        \multicolumn{4}{c}{Models} &
        \multirow{2}{*}{Human} \\
        \cmidrule(lr){3-6}
        & & \texttt{llama-4 -17b-16e} & \texttt{claude-4-5 -sonnet} & \texttt{gpt-5 -mini} & \texttt{gemini-3 -pro} & \\
        \midrule
        \multirow{3}{*}{Accuracy}
            & Material              & 27.0\% & 84.5\% & 92.7\% & 70.4\% & 97.8\% \\
            & Shape                 & 22.2\% & 55.8\% & 53.4\% & 50.3\% & 99.2\% \\
            & Size                  &  3.2\% & 68.1\% & 77.0\% & 52.8\% & 91.7\% \\
        \midrule
        \multirow{2}{*}{Safety}
            & Physical Plausibility & 77.8\% & 85.0\% & 68.3\% & 92.0\% & 98.3\% \\
            & Structural Stability  & 25.0\% & 57.1\% & 50.7\% & 42.7\% & 93.9\% \\
        \midrule
        \multirow{2}{*}{Planning}
            & Efficiency            & 47.1\% & 56.2\% & 55.6\% & 59.7\% & 95.6\% \\
            & Dependency            &  7.4\% & 96.3\% & 68.6\% & 78.2\% & 90.0\% \\
        \bottomrule
    \end{tabularx}
    \caption{
\textbf{Performance on MineCEraft across different model families and sizes.}
Human results are reported on MineCEraft-Lite for reference (see Appendix~\ref{sec:human_eval}).
Overall, LLMs underperform human participants across most evaluation categories.
For Lite-only comparisons across all models, see Table~\ref{tab:performance_lite_full} in Appendix~\ref{sec:rep_lite}.
For evaluation results on more models, see Table~\ref{tab:performance_many_models} in Appendix~\ref{sec:rep_lite}.
}
\label{tab:performance}
\end{table*}

%% file: sections/abc.tex

\input{figs/qualitative_abc}

\section{Toy Example: Arch Bridge Challenge} \label{sec:abc}
As a point of departure, we begin with a very simple experiment, the \emph{Arch Bridge Challenge} (ABC). 
Consider the following prompt:
\begin{quote}
\begin{Verbatim}[breaklines,breakanywhere]
Let's build an arch bridge viewed from the side using stones (#) and empty spaces (-).
\end{Verbatim}
\end{quote}
Chances are that even a state-of-the-art LLM (e.g., ChatGPT 5)  will generate a non-arch structure (or not even a bridge at all), as illustrated in Figure~\ref{fig:qualitative_abc}.
In contrast, also consider an LLM output when prompted with: 
\begin{quote}
\begin{Verbatim}[breaklines,breakanywhere]
Draw an arch bridge as an image.
\end{Verbatim}
\end{quote}
In general, with the second prompt, a text-to-image model can produce a beautiful, well-formed bridge. 
This contrast suggests that certain forms of knowledge within LLMs are not internally well-connected; in other words, their ability to generalize across modalities is limited.

Put differently, although an LLM may respond as if it understands what an arch is, merely changing the output modality from image generation to text generation reveals a failure of generalization. 
This example, therefore, can serve to debunk the illusion that LLMs truly \textit{understand} the structural principles of an arch bridge.

%% file: figs/qualitative_abc.tex
\begin{figure*}[t]
\centering
\setlength{\tabcolsep}{15pt}
\renewcommand{\arraystretch}{1.0}

\begin{tabular}{ccc}

\begin{minipage}[t]{0.28\textwidth}
\centering
\includegraphics[width=\linewidth]{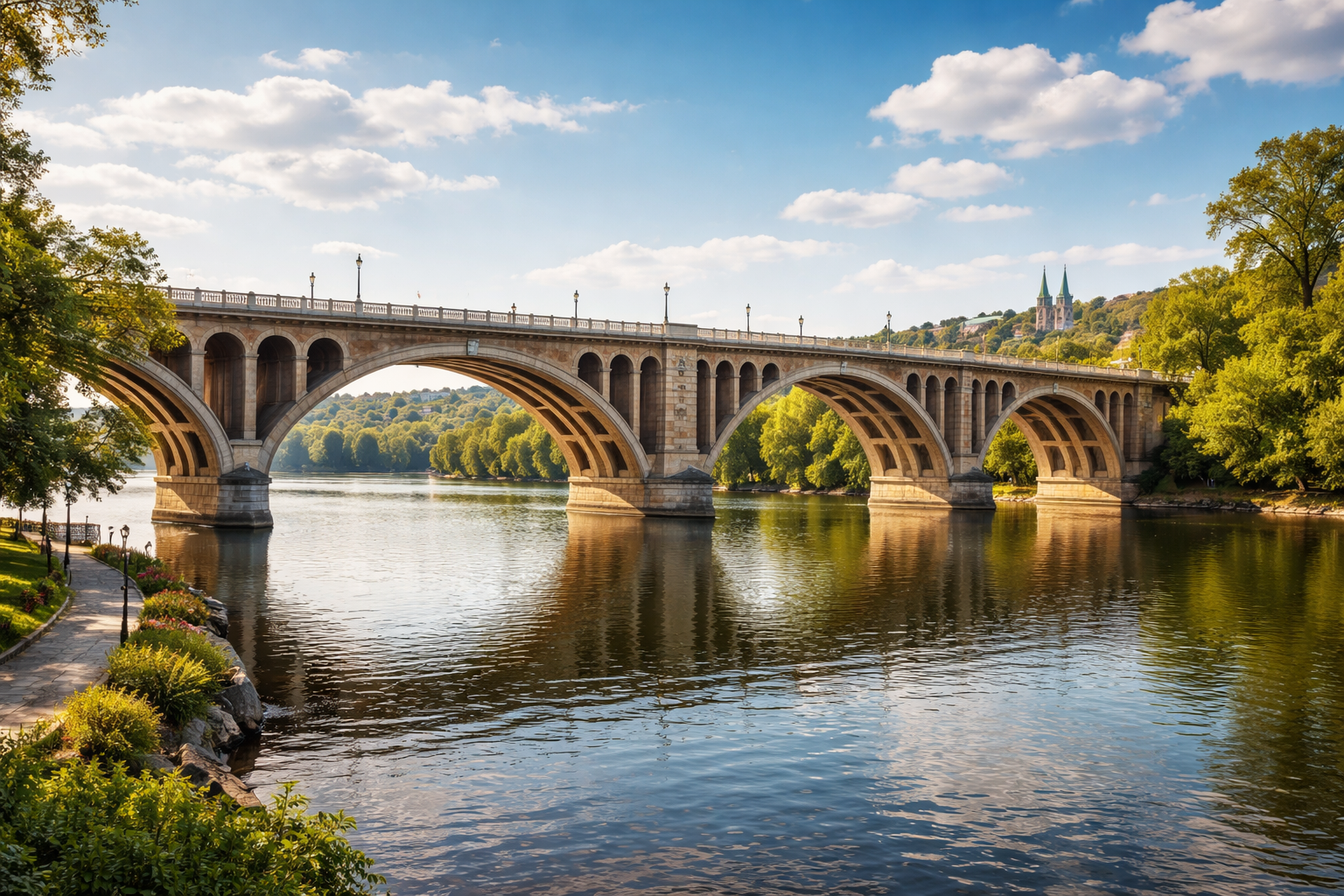}
\end{minipage}
&
\begin{minipage}[t]{0.28\textwidth}
\centering
\includegraphics[width=\linewidth]{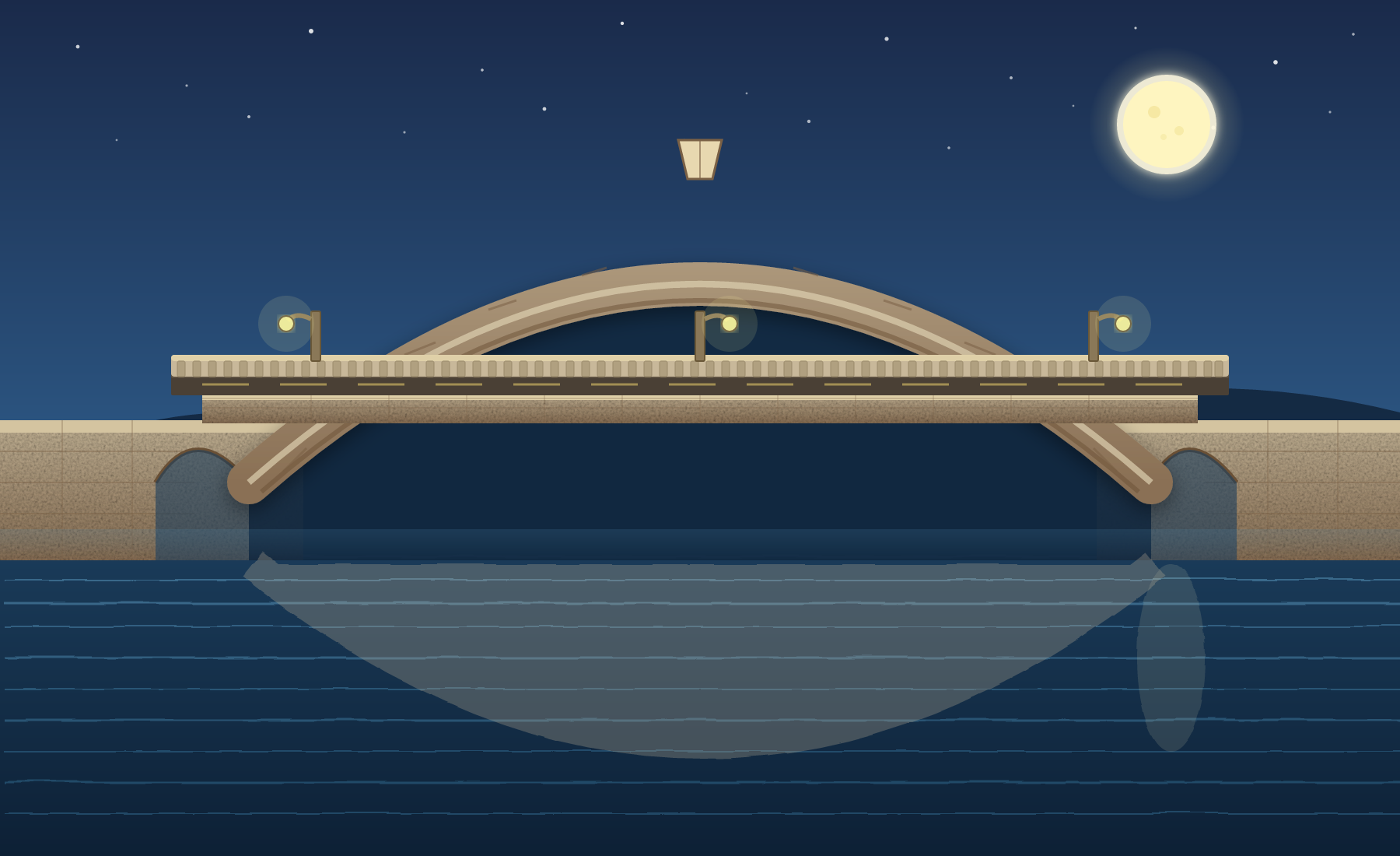}
\end{minipage}
&
\begin{minipage}[t]{0.28\textwidth}
\centering
\includegraphics[width=\linewidth]{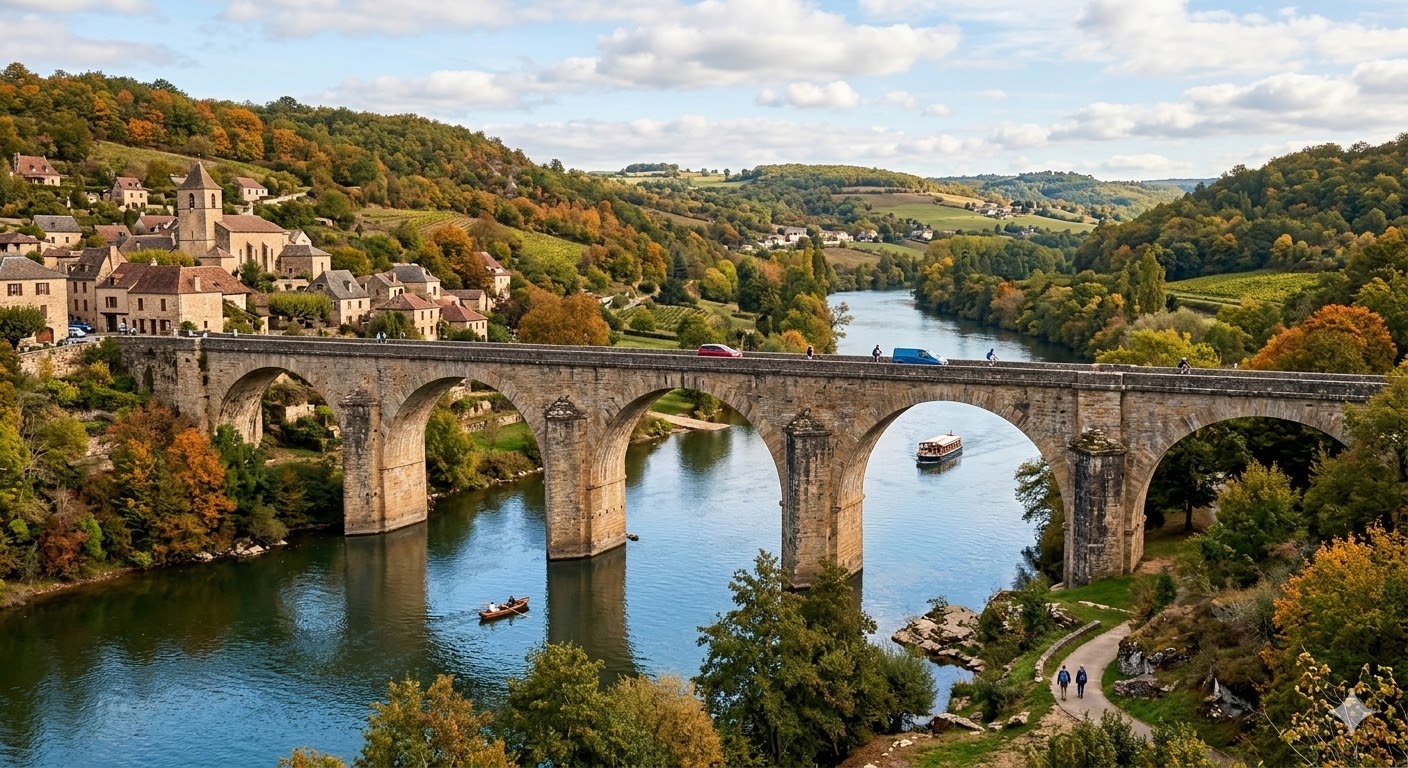}
\end{minipage}
\\[6pt]

\begin{minipage}[t]{0.28\textwidth}
\centering
\small
(a) ChatGPT 5.4 Thinking (Image)
\end{minipage}
&
\begin{minipage}[t]{0.28\textwidth}
\centering
\small
(b) Claude Sonnet 4.6 Extended (Image)
\end{minipage}
&
\begin{minipage}[t]{0.28\textwidth}
\centering
\small
(c) Gemini 3 Thinking (Image)
\end{minipage}
\\[32pt]

\begin{minipage}[t]{0.28\textwidth}
\centering
\begin{BVerbatim}[fontsize=\tiny]
-----------------#-----------------
---------------#####---------------
-------------#########-------------
-----------#############-----------
---------#################---------
-------#####################-------
-----#########################-----
---#############################---
-#################################-
###################################
\end{BVerbatim}
\end{minipage}
&
\begin{minipage}[t]{0.28\textwidth}
\centering
\begin{BVerbatim}[fontsize=\tiny]
-----------#######-----------
----------#-------#----------
---------#---------#---------
--------#-----------#--------
-------#-------------#-------
------#---------------#------
-----#-----------------#-----
----##-------------------##--
---#-------------------------#
--###########################--
--###########################--
##############################
\end{BVerbatim}
\end{minipage}
&
\begin{minipage}[t]{0.28\textwidth}
\centering
\begin{BVerbatim}[fontsize=\tiny]
      [ Roadway ]
#######################
#######---------#######
#####-------------#####
####---------------####
###-----------------###
###-----------------###
###-----------------###
\end{BVerbatim}
\end{minipage}
\\[6pt]

\begin{minipage}[t]{0.28\textwidth}
\centering
\begin{BVerbatim}[fontsize=\tiny]
----------------#############----------------
-------------####---------####-------------
----------####---------------####----------
--------###---------------------###--------
------###-------------------------###------
-----##---------------------------##-----
----##-----------------------------##----
----##-----------------------------##----
-----##---------------------------##-----
------###-----------------------###------
--------####-----------------####--------
-----------#####---------#####-----------
---------------#########---------------
\end{BVerbatim}
\end{minipage}
&
\begin{minipage}[t]{0.28\textwidth}
\centering
\begin{BVerbatim}[fontsize=\fontsize{3}{3}\selectfont]
###############################################################################
###---------------------------------------------------------------------------###
##-----------------------------------------------------------------------------##
#-------------------------------------------------------------------------------#
#---------------------------------########---------------------------------#
#-----------------------------################-----------------------------#
#---------------------------####################---------------------------#
#-------------------------########################-------------------------#
##-----------------------############################-----------------------##
####-------------------################################-------------------####
######-----------------##################################-----------------######
########-------------######################################-------------########
##########-----------########################################-----------##########
###############################################################################
###############################################################################
~~~~~~~~~~~~~~~~~~~~~~~~~~~~~~~~~~~~~~~~~~~~~~~~~~~~~~~~~~~~~~~~~~~~~~~~~~~~~~~
~~~~~~~~~~~~~~~~~~~~~~~~~~~~~~~~~~~~~~~~~~~~~~~~~~~~~~~~~~~~~~~~~~~~~~~~~~~~~~~
\end{BVerbatim}
\end{minipage}
&
\begin{minipage}[t]{0.28\textwidth}
\centering
\begin{BVerbatim}[fontsize=\tiny]
      _______________________
     /       Roadway         \
    ###########################
    ###########################
    #######-------------#######
    #####-----------------#####
    ####-------------------####
    ###---------------------###
    ##-----------------------##
    ##-----------------------##
\end{BVerbatim}
\end{minipage}
\\[6pt]

\begin{minipage}[t]{0.28\textwidth}
\centering
\begin{BVerbatim}[fontsize=\tiny]
----------------#############----------------
-------------###-------------###-------------
----------###-------------------###----------
--------##------------------------##---------
------##----------------------------##-------
-----##------------------------------##------
----##--------------------------------##-----
####----------------------------------####---
####----------------------------------####---
####----------------------------------####---
\end{BVerbatim}
\end{minipage}
&
\begin{minipage}[t]{0.28\textwidth}
\centering
\begin{BVerbatim}[fontsize=\tiny]
-----------------------------------------------
--------------------#####----------------------
-----------------###-----###-------------------
---------------##-----------##-----------------
-------------##---------------##---------------
-----------##-------------------##-------------
----------#-----------------------#------------
##########---------------------------##########
##########---------------------------##########
##########~~~~~~~~~~~~~~~~~~~~~~~~~~~##########
###############~~~~~~~~~~~~~~~~~~~############
###############~~~~~~~~~~~~~~~~~~~############
###############~~~~~~~~~~~~~~~~~~~############
\end{BVerbatim}
\end{minipage}
&
\begin{minipage}[t]{0.28\textwidth}
\centering
\begin{BVerbatim}[fontsize=\fontsize{3}{3}\selectfont]
      _________________________ ROADWAY _________________________
     |                                                           |
     #############################################################
     #############################################################
     ###########---------- STONE STRUCTURE ----------###########
     ########-------------------------------------------########
     ######-----------------------------------------------######
     #####------------------- OPEN ARCH -------------------#####
     ####---------------------------------------------------####
     ###-----------------------------------------------------###
     ###-----------------------------------------------------###
     ###-----------------------------------------------------###
    ####-----------------------------------------------------####
   ######---------------------------------------------------######
  ########-------------------------------------------------########
 ##########-----------------------------------------------##########
###########-----------------------------------------------###########
\end{BVerbatim}
\end{minipage}
\\[6pt]

\begin{minipage}[t]{0.28\textwidth}
\centering
\begin{BVerbatim}[fontsize=\fontsize{5}{5}\selectfont]
----------------------####----------------------
-------------------##########-------------------
-----------------##############-----------------
---------------##################---------------
-------------######################-------------
-----------##########################-----------
---------##########------##########---------
-------##########----------##########-------
#####-##########--------------##########-#####
##############################################
##############################################
\end{BVerbatim}
\end{minipage}
&
\begin{minipage}[t]{0.28\textwidth}
\centering
\begin{BVerbatim}[fontsize=\tiny]
--------------------#########--------------------
-------------------###########-------------------
------------------#############------------------
-----------------###############-----------------
################-###############-################
################--#############--################
################---###########---################
################-----#######-----################
################---------#-------################
################-----------------################
#################################################
#################################################
#################################################
\end{BVerbatim}
\end{minipage}
&
\begin{minipage}[t]{0.28\textwidth}
\centering
\begin{BVerbatim}[fontsize=\tiny]
      ___________________________
     /                           \
    |   #########################   |  <-- Roadway
    |   #####-------------#######   |
    |   ###-----------------#####   |
    |   ##-------------------####   |  <-- The Arch
    |   ##-------------------####   |
    |   ##-------------------####   |  <-- Abutments
    |   #########################   |
    |_______________________________|
\end{BVerbatim}
\end{minipage}
\\[6pt]

\begin{minipage}[t]{0.28\textwidth}
\centering
\small
(d) ChatGPT 5.4 Thinking
\end{minipage}
&
\begin{minipage}[t]{0.28\textwidth}
\centering
\small
(e) Claude Sonnet 4.6 Extended
\end{minipage}
&
\begin{minipage}[t]{0.28\textwidth}
\centering
\small
(f) Gemini 3 Thinking
\end{minipage}
\\

\end{tabular}

\caption{
\textbf{Qualitative examples of building arch bridges in 2D}. While LLMs can generate arch bridges when prompted with \textit{“Draw an arch bridge as an image.”}, they often produce irregular or structurally incorrect shapes when asked in a text-based format, such as \textit{“Let's build an arch bridge viewed from the side using stones (\#) and empty spaces (-).”}. By comparison, Gemini 3 can generate a relatively arch-like structure even in the text-based setting.
}
\label{fig:qualitative_abc}
\end{figure*}

%% file: sections/algo.tex
\section{Evaluation Algorithms}
\label{sec:algo}

\paragraph{Notation and Terminologies.}
We represent a structure as a block set $B$, where each block
$b \in B$ has integer coordinates $(x,y,z)$ and a material label
$b.\mathrm{material}$. When construction order is relevant,
we write $B=(b_1,\dots,b_n)$ for the ordered block-placement
sequence.
We use $\mathrm{match}(b.\mathrm{material}, m)$ to denote
material matching under the canonical material taxonomy used
in the benchmark.
Some checks operate on predefined subsets of blocks.
These are specified using a region selector $r$ that extracts
a subset $R = r(B) \subseteq B$ (e.g., corners at a given
height, boundary walls, or the top layer).
For connectivity in 3D we use 6-neighbor face adjacency.
For planar flood-fill operations on the $x$--$z$ grid
(e.g., room counting), we use 8-neighborhood connectivity.

Concretely, metrics are evaluated by the following algorithms:
\begin{itemize}
    \item Material accuracy: Algorithms~\ref{alg:mat-global}--\ref{alg:mat-region}.
    \item Shape accuracy: Algorithms~\ref{alg:concavity}--\ref{alg:stairs}.
    \item Size accuracy: Algorithms~\ref{alg:size-box} and \ref{alg:size-axis}.
    \item Physical plausibility: Algorithm~\ref{alg:ground-connectivity}.
    \item Structural stability: Appendix~\ref{app:von_mises}.
    \item Efficiency: Algorithm~\ref{alg:path-efficiency}.
    \item Dependency: Algorithm~\ref{alg:material-order}.
\end{itemize}

\begin{algorithm}[t]
\caption{Global Material Uniformity}
\label{alg:mat-global}
\begin{algorithmic}[1]
\Require Block set $B$, target material $m$
\Ensure \textsc{True} iff every block uses material $m$
\If{$B = \emptyset$}
    \State \Return \textsc{False}
\EndIf
\ForAll{$b \in B$}
    \If{$\neg \mathrm{match}(b.\mathrm{material}, m)$}
        \State \Return \textsc{False}
    \EndIf
\EndFor
\State \Return \textsc{True}
\end{algorithmic}
\end{algorithm}

\begin{algorithm}[t]
\caption{Counting Blocks of a Given Material}
\label{alg:material-count}
\begin{algorithmic}[1]
\Require Block set $B$, expected material $m$ (optional), substring flag $s$
\Ensure Number of blocks whose material matches $m$

\State $count \gets 0$

\For{each block $b \in B$}
    \If{$m$ is \texttt{None}}
        \State $count \gets count + 1$
    \Else
        \State let $mat$ be the material of $b$
        \If{$s$ is \textsc{True}}
            \If{$m$ is a substring of $mat$}
                \State $count \gets count + 1$
            \EndIf
        \Else
            \If{$mat = m$}
                \State $count \gets count + 1$
            \EndIf
        \EndIf
    \EndIf
\EndFor

\State \Return $count$
\end{algorithmic}
\end{algorithm}

\begin{algorithm}[t]
\caption{Regional Material Ratio Check}
\label{alg:mat-region}
\begin{algorithmic}[1]
\Require Block set $B$, region selector $r$, target material $m$, threshold $\tau$
\Ensure \textsc{True} iff the selected region mostly uses material $m$
\State $R \gets r(B)$
\If{$R = \emptyset$}
    \State \Return \textsc{False}
\EndIf
\State $c \gets |\{ b \in R : \mathrm{match}(b.\mathrm{material}, m) \}|$
\State \Return $(c / |R|) \ge \tau$
\end{algorithmic}
\end{algorithm}

\begin{algorithm}[t]
\caption{Surface Concavity Check}
\label{alg:concavity}
\begin{algorithmic}[1]
\Require Block set $B$, mode $m \in \{\textsc{Top}, \textsc{Bottom}\}$
\Ensure \textsc{True} iff some axis-aligned slice is concave
\If{$B = \emptyset$}
    \State \Return \textsc{True}
\EndIf
\State Compute surface height map $S(x,z)$ using
\Statex \hspace{1em} $\max_y$ for \textsc{Top} and $\min_y$ for \textsc{Bottom}
\State Construct all 1D slices of $S$ by fixing $x$ or $z$
\ForAll{slices $P$}
    \If{$P$ satisfies the chord-based concavity condition}
        \State \Return \textsc{True}
    \EndIf
\EndFor
\State \Return \textsc{False}
\end{algorithmic}
\end{algorithm}

\begin{algorithm}[t]
\caption{Room Counting}
\label{alg:rooms}
\begin{algorithmic}[1]
\Require Block set $B$, walking height $y$ (default $y=1$), target room count $k$
\Ensure \textsc{True} iff at least $k$ enclosed rooms exist

\State Let $O$ be the set of occupied cells $(x,z)$ on the $x$--$z$ plane at height $y$
\If{$O = \emptyset$}
    \State \Return \textsc{False}
\EndIf

\State Compute the bounding box of $O$ and expand it by one cell in all directions
\State Initialize a queue with all boundary cells of the expanded box that are not in $O$
\State Let $A$ be the set of cells reachable from these boundary cells

\Comment{Outside air detection using the Flood-fill algorithm}
\While{the queue is not empty}
    \State Pop a cell $(x,z)$
    \For{each of its 8-neighbor cells $(x',z')$}
        \If{$(x',z')$ is inside the bounding box and not in $O$ and not in $A$}
            \State Add $(x',z')$ to $A$ and push it into the queue
        \EndIf
    \EndFor
\EndWhile

\State Count connected components of empty cells that are neither in $O$ nor in $A$
\If{the number of such enclosed components $\ge k$}
    \State \Return \textsc{True}
\Else
    \State \Return \textsc{False}
\EndIf
\end{algorithmic}
\end{algorithm}

\begin{algorithm}[t]
\caption{Minimum Inter-Cluster Span Check}
\label{alg:min-span}
\begin{algorithmic}[1]
\Require Block set $B$, height $y$, minimum span threshold $s_{\min}$, neighborhood option \texttt{use\_8\_neighbors}
\Ensure \textsc{True} iff every pair of clusters at height $y$ is at least $s_{\min}$ apart

\State Extract all occupied cells on the $x$--$z$ plane at height $y$
\State Compute connected components $C_1, C_2, \dots, C_n$ on this plane
\Statex \hspace{\algorithmicindent} using 4-neighborhood by default, or 8-neighborhood if \texttt{use\_8\_neighbors} is enabled

\If{$n < 2$}
    \State \Return \textsc{False}
\EndIf

\For{each unordered pair of distinct clusters $(C_i, C_j)$}
    \State Compute the minimum pairwise Euclidean distance
    \[
    d(C_i, C_j)=
    \min_{\substack{p\in C_i\\ q\in C_j}}
    \lVert p-q\rVert_2
    \]
    \If{$d(C_i, C_j) < s_{\min}$}
        \State \Return \textsc{False}
    \EndIf
\EndFor

\State \Return \textsc{True}
\end{algorithmic}
\end{algorithm}

\begin{algorithm}[t]
\caption{Flat Exit Path}
\label{alg:flat-path}
\begin{algorithmic}[1]
\Require Block set $B$, optional start selector $m_s$
\Ensure $1$ iff a flat path to the outside exists from every start

\If{$B=\emptyset$} \State \Return $1$ \EndIf

\State $y_f \gets \min_y(B)$ \Comment{floor level}
\If{$y_f > -1$} \State \Return $0$ \EndIf

\State $F \gets$ walkable cells at level $y_f$
\State $P \gets$ start positions (matching $m_s$, or centroid if none)

\ForAll{$p \in P$}
    \If{$y(p) \neq y_f + 1$} \State \Return $0$ \EndIf

    \State Run BFS on the $x$--$z$ plane from $p$
    \Statex \hspace{\algorithmicindent} allowing moves only within $F$ 
    \Statex \hspace{\algorithmicindent} with sufficient head clearance

    \If{no path reaches outside the bounding box}
        \State \Return $0$
    \EndIf
\EndFor

\State \Return $1$
\end{algorithmic}
\end{algorithm}

\begin{algorithm}[t]
\caption{Reachability by Stairs}
\label{alg:stairs}
\begin{algorithmic}[1]
\Require Block set $B$, target level $y_{\min}$
\Ensure $1$ iff an agent can reach height $\ge y_{\min}$
\State Initialize BFS from all ground blocks with head clearance
\While{queue not empty}
    \State Pop $(x,y,z)$
    \If{$y \ge y_{\min}$}
        \State \Return $1$
    \EndIf
    \State Enqueue valid 4-neighbor moves at level $y$ or $y+1$
\EndWhile
\State \Return $0$
\end{algorithmic}
\end{algorithm}

\begin{algorithm}[t]
\caption{Bounding-Box Size Check}
\label{alg:size-box}
\begin{algorithmic}[1]
\Require Block set $B$, planar size $(s_x,s_z)$, optional height $s_y$, mode $m \in \{\textsc{Equal},\textsc{AtMost}\}$
\Ensure $1$ iff the bounding-box size constraint holds
\If{$B = \emptyset$}
    \State \Return $0$
\EndIf
\State Compute bounding-box ranges $(d_x,d_y,d_z)$
\State $(r_x,r_z) \gets (s_x-1, s_z-1)$
\If{$m = \textsc{Equal}$}
    \State \Return $1$ iff $\{d_x,d_z\} = \{r_x,r_z\}$ and (if given) $d_y = s_y - 1$
\Else
    \State \Return $1$ iff $\{d_x,d_z\} \le \{r_x,r_z\}$ and (if given) $d_y \le s_y - 1$
\EndIf
\end{algorithmic}
\end{algorithm}

\begin{algorithm}[t]
\caption{Vertical and Minimum-Size Check}
\label{alg:size-axis}
\begin{algorithmic}[1]
\Require Block set $B$, constraint $c$
\Ensure $1$ iff the size constraint holds
\If{$B = \emptyset$}
    \State \Return $0$
\EndIf
\State Compute $\min_y(B)$, $\max_y(B)$, and $(d_x,d_y,d_z)$
\If{$c$ specifies exact minimum depth}
    \State \Return $1$ iff $\min_y(B) = y^\star$
\ElsIf{$c$ specifies minimum depth}
    \State \Return $1$ iff $\min_y(B) \le y^\star$
\ElsIf{$c$ specifies minimum height}
    \State \Return $1$ iff $\max_y(B) \ge y^\star$
\ElsIf{$c$ specifies minimum size}
    \State \Return $1$ iff $d_x,d_y,d_z \ge s^\star - 1$
\EndIf
\State \Return $0$
\end{algorithmic}
\end{algorithm}

\begin{algorithm}[t]
\caption{Ground Connectivity Check}
\label{alg:ground-connectivity}
\begin{algorithmic}[1]
\Require Block set $B$
\Ensure $1$ iff every block connects to the ground
\If{$B = \emptyset$}
    \State \Return $1$
\EndIf
\State $G \gets \{ b \in B : y(b) \le 0 \}$
\If{$G = \emptyset$}
    \State \Return $0$
\EndIf
\State Run BFS from all blocks in $G$ using 6-neighbor adjacency
\If{all blocks in $B$ are visited}
    \State \Return $1$
\Else
    \State \Return $0$
\EndIf
\end{algorithmic}
\end{algorithm}

\begin{algorithm}[t]
\caption{Path Efficiency}
\label{alg:path-efficiency}
\begin{algorithmic}[1]
\Require Ordered block sequence $B=(b_1,\dots,b_n)$
\Ensure Efficiency score in $[0,1]$
\If{$n \le 1$}
    \State \Return $1$
\EndIf
\State $L \gets \sum_{i=2}^{n} \|b_i - b_{i-1}\|_1$
\State $L^\star \gets n - 1$
\If{$L \le 0$}
    \State \Return $1$
\EndIf
\State \Return $\min(1, L^\star / L)$
\end{algorithmic}
\end{algorithm}

\begin{algorithm}[t]
\caption{Material Sequence Order Check}
\label{alg:material-order}
\begin{algorithmic}[1]
\Require Ordered block sequence $B=(b_1,\dots,b_n)$, target sequence $M=(m_1,\dots,m_k)$
\Ensure $1$ iff $M$ appears in order in $B$
\If{$M = \emptyset$}
    \State \Return $1$
\EndIf
\If{$B = \emptyset$}
    \State \Return $0$
\EndIf
\State $j \gets 1$
\For{$i=1$ to $n$}
    \If{$\mathrm{match}(b_i.\mathrm{material}, m_j)$}
        \State $j \gets j+1$
        \If{$j > k$}
            \State \Return $1$
        \EndIf
    \EndIf
\EndFor
\State \Return $0$
\end{algorithmic}
\end{algorithm}

%% file: sections/additional.tex
\section{Additional Results and Statistics}
\label{sec:additional}

\subsection{Overall Problem Accuracy}
\label{sec:overall_acc}

One may ask the following two questions. 
First, at the level of the entire criteria of each problem, how often does a model satisfy all binary accuracy-related requirements? 
Second, among the constructions that do satisfy those binary requirements, how high are the graded/continuous metrics, such as efficiency and structural stability?

These questions are particularly important in construction, where even a single mistake is often not allowed. 
Since simply averaging heterogeneous metrics can obscure risk, a stricter and more reliable aggregation at the problem level is reported in Table~\ref{overall_acc}. 
We first define a construction as \textit{accurate} if it satisfies the intersection of \textit{all} accuracy criteria. 
We then further aggregate across safety and planning by additionally requiring physical plausibility and dependency satisfaction, which are also binary, producing an overall success criterion that reflects the intersection of these dimensions.
As shown in Table~\ref{overall_acc}, the resulting problem-level success rates are substantially lower than the instruction-level accuracies reported earlier. 
This stricter aggregation makes it clearer that current LLM-based agents still struggle to perform construction tasks in a reliable manner. 

A second question is whether constructions that are \textit{accurate} in this strict binary sense tend to achieve better performance on graded metrics. 
Table~\ref{acc_with_graded} suggests two main observations (excluding \texttt{llama-4-17b-16e}, which produced no fully accurate constructions on the relevant subsets). 
First, binary accuracy does not necessarily translate into improved efficiency. 
That is, even when a model produces a construction that correctly satisfies the required material, shape, and size constraints, the resulting building process is not consistently more movement-efficient.

Second, constructions that satisfy the accuracy criteria sometimes exhibit slightly higher structural stability for some models (\texttt{gpt5-mini} and \texttt{gemini-3-pro}). 
This observation is consistent with basic engineering intuition. 
There is no inherent reason to expect that a structure built exactly according to specification must also have been constructed efficiently. 
In contrast, constructions that satisfy material, shape, and size constraints tend to exhibit slightly higher structural stability.

\input{tables/overall_acc}
\input{tables/acc_with_graded}

\subsection{Effect of Chain-of-Thought Prompting}
We further analyze the effect of chain-of-thought style prompting on construction performance. While prompts such as ``Let's think step by step'' have been shown to improve reasoning performance in domains such as mathematics \citep{kojima2022large}, we observe little to no measurable improvement in embodied construction tasks. In contrast, prompts that explicitly encourage reasoning about construction order or movement planning lead to modest improvements in planning-related metrics. Detailed results are reported in Table~\ref{tab:construction_efficiency}.

\begin{table}[t]
\centering
\small
\renewcommand{\arraystretch}{1.05}
\begin{tabularx}{\linewidth}{l*{4}{>{\centering\arraybackslash}X}}
\toprule
Prompt Type
& \texttt{llama -4 -scout}
& \texttt{claude -4-5 -sonnet}
& \texttt{gpt-5 -mini}
& \texttt{gemini -3-pro} \\
\midrule
Baseline
& \textbf{54.8}\% & 58.3\% & 58.3\% & 58.3\% \\

Step-by-step
& 52.4\% & 58.3\% & 58.3\% & 58.3\% \\
$\Delta$ (vs.\ Baseline)
& -2.4\% & 0.0\% & 0.0\% & 0.0\% \\

Order-aware
& 52.4\% & \textbf{79.2\%} & \textbf{79.2\%} & \textbf{100.0\%} \\
$\Delta$ (vs.\ Baseline)
& -2.4\% & {+20.9\%} & {+20.9\%} & {+41.7\%} \\
\bottomrule
\end{tabularx}
\caption{Construction efficiency (\%, higher is better) for the foundation-building task under different prompting strategies. The baseline prompt contains only the task instruction. \textbf{Step-by-step} adds the generic reasoning cue \textit{``Let's think step by step.''}, whereas \textbf{Order-aware} adds the task-specific cue \textit{``Let's think about the order in which we should place the blocks to optimize the movement path.''} Rows labeled $\Delta$ show the change in efficiency relative to the baseline. The generic reasoning cue does not improve efficiency, while the order-aware cue yields substantial gains for several models.}
\label{tab:construction_efficiency}
\end{table}

\subsection{Sensitivity to Instruction Verbs}
We examine how variations in instruction wording affect model performance. Specifically, we compare prompts that differ only in the verb used (e.g., ``build'', ``create'', or ``construct''). Despite conveying similar semantic intent, we observe measurable performance differences across several models and evaluation criteria, as in Table~\ref{tab:verbs}. This suggests that LLM behavior on construction tasks can be sensitive to subtle variations in natural-language instructions.

\begin{table}[t]
\centering
\small
\setlength{\tabcolsep}{4pt}

\begin{tabular}{lcc}
\hline
 & \textbf{Professional abb.} & \textbf{Plain words} \\
\hline
Material & \textbf{66.7} & 50.0 \\
Shape & \textbf{66.7} & \textbf{66.7} \\
Physical Plausibility & \textbf{100.0} & \textbf{100.0} \\
Structural Stability & \textbf{66.7} & 16.7 \\
Dependency & 66.7 & \textbf{100.0} \\
\hline
\end{tabular}

\vspace{4pt}

\begin{tabular}{lccc}
\hline
 & \textbf{``build''} & \textbf{``construct''} & \textbf{``create''} \\
\hline
Material & 50.0 & \textbf{100.0} & 25.0 \\
Shape & 50.0 & \textbf{100.0} & 50.0 \\
Dependency & 75.0 & \textbf{100.0} & 75.0 \\
Physical Plausibility & \textbf{100.0} & \textbf{100.0} & \textbf{100.0} \\
Structural Stability & 25.0 & \textbf{50.0} & \textbf{50.0} \\
\hline
\end{tabular}

\caption{Sensitivity to instruction wording for \texttt{claude-4-5-sonnet} evaluated on 72 construction tasks of roofs and columns in Section~\ref{sec:se}. 
Top: comparison between a field-specific abbreviation (SOMD) and its plain-language equivalent (“slab on metal deck”) shows little difference in performance, suggesting that construction-specific terminology is reasonably well captured. 
Bottom: prompts differing only in the verb (“build”, “construct”, “create”) show measurable differences.}
\label{tab:verbs}
\end{table}

\subsection{Sensitivity to System Prompt}
\label{sec:system_prompt_sensitivity}

We further examine whether the system prompt affects model performance.
In particular, we remove the instruction ``Be very brief in your responses''
from the original MINDcraft-based system prompt \citep{mindcraft2025} while keeping the remaining
setup unchanged. Table~\ref{tab:system_prompt_sensitivity} compares the
original prompt with this modified prompt for four representative models.

Removing the brevity constraint substantially improves dependency planning
across all four models, with gains ranging from 20.0 to 40.0 percentage
points. However, the effect is not uniformly positive across other criteria.
For example, some models improve substantially in structural stability,
whereas shape, physical plausibility, or efficiency can decrease.
These results indicate that construction performance can be sensitive to
system-prompt design, motivating a more systematic study of prompt design
and optimization in future work.

\begin{table*}[t]
    \centering
    \scriptsize
    \setlength{\tabcolsep}{3.5pt}
    \renewcommand{\arraystretch}{1.08}

    \resizebox{\textwidth}{!}{%
    \begin{tabular}{llrrrrrrr}
        \toprule
        Model & System Prompt
        & Material & Shape & Size
        & P.P. & S.S. & Eff. & Dep. \\
        \midrule

        \multirow{3}{*}{\texttt{llama-4-scout-17b-16e}}
        & Original
        & 20.6 & 23.8 & 24.6 & 85.7 & 16.7 & 37.8 & 0.0 \\
        & w/o brevity constraint
        & 33.3 & 29.5 & 16.7 & 83.3 & 0.0 & 84.2 & 20.0 \\
        & $\Delta$
        & +12.7 & +5.7 & -7.9 & -2.4 & -16.7 & +46.4 & +20.0 \\
        \midrule

        \multirow{3}{*}{\texttt{claude-4-5-sonnet}}
        & Original
        & 84.1 & 60.5 & 83.3 & 83.3 & 55.6 & 54.9 & 60.0 \\
        & w/o brevity constraint
        & 90.5 & 47.5 & 91.7 & 66.7 & 91.9 & 53.1 & 100.0 \\
        & $\Delta$
        & +6.4 & -13.0 & +8.4 & -16.6 & +36.3 & -1.8 & +40.0 \\
        \midrule

        \multirow{3}{*}{\texttt{gpt-5-mini}}
        & Original
        & 87.1 & 52.5 & 83.3 & 59.2 & 67.1 & 54.9 & 62.2 \\
        & w/o brevity constraint
        & 95.2 & 45.9 & 75.0 & 60.0 & 72.4 & 53.1 & 100.0 \\
        & $\Delta$
        & +8.1 & -6.6 & -8.3 & +0.8 & +5.3 & -1.8 & +37.8 \\
        \midrule

        \multirow{3}{*}{\texttt{gemini-3-pro}}
        & Original
        & 71.4 & 47.5 & 38.0 & 97.8 & 49.4 & 69.8 & 60.0 \\
        & w/o brevity constraint
        & 81.0 & 65.6 & 50.0 & 80.0 & 79.4 & 53.1 & 100.0 \\
        & $\Delta$
        & +9.6 & +18.1 & +12.0 & -17.8 & +30.0 & -16.7 & +40.0 \\

        \bottomrule
    \end{tabular}%
    }

    \caption{
    Sensitivity to the brevity constraint in the system prompt.
    We compare the original MINDcraft-based system prompt with a variant
    that removes the instruction ``Be very brief in your responses.''
    Values are percentages, and $\Delta$ denotes the change relative to
    the original prompt.
    }
    \label{tab:system_prompt_sensitivity}
\end{table*}

\subsection{Additional Multimodal Model Comparison}

To further investigate the potential of multimodal models, we additionally
compare GPT-4 with GPT-4o. As shown in
Table~\ref{tab:gpt4_gpt4o}, GPT-4o outperforms GPT-4 across all
evaluation criteria, with particularly large improvements in material
accuracy and structural stability. Although this comparison does not
directly evaluate visual perception, as visual observations are not provided
during evaluation, the result provides indirect evidence that capabilities
of multimodal models may still benefit embodied construction in a text-based
interaction setting.

\begin{table}[t]
    \centering
    \small
    \setlength{\tabcolsep}{4pt}
    \renewcommand{\arraystretch}{1.05}
    \resizebox{\columnwidth}{!}{%
    \begin{tabular}{lccccccc}
        \toprule
        Model & Material & Shape & Size & P.P. & S.S. & Eff. & Dep. \\
        \midrule
        GPT-4
        & 42.9\% & 50.8\% & 50.0\% & 70.0\% & 0.0\% & 53.1\% & 40.0\% \\
        GPT-4o
        & 85.7\% & 55.9\% & 61.1\% & 84.4\% & 67.8\% & 62.0\% & 60.0\% \\
        \bottomrule
    \end{tabular}%
    }
    \caption{Additional GPT-4 vs.\ GPT-4o comparison.}
    \label{tab:gpt4_gpt4o}
\end{table}

%% file: tables/overall_acc.tex
\begin{table*}[t]
\centering
\small
\setlength{\tabcolsep}{5pt}
\renewcommand{\arraystretch}{1.15}
\resizebox{\linewidth}{!}{
\begin{tabular}{l c cc cc cc cc}
\toprule
& & \multicolumn{2}{c}{\texttt{llama-4-17b-16e}} 
  & \multicolumn{2}{c}{\texttt{claude-4-5-sonnet}} 
  & \multicolumn{2}{c}{\texttt{gpt5-mini}} 
  & \multicolumn{2}{c}{\texttt{gemini-3-pro}} \\
\cmidrule(lr){3-4}\cmidrule(lr){5-6}\cmidrule(lr){7-8}\cmidrule(lr){9-10}
Metric & \#Problems 
& \#Correct & Accuracy 
& \#Correct & Accuracy 
& \#Correct & Accuracy 
& \#Correct & Accuracy \\
\midrule
\textbf{Overall Accuracy {\scriptsize ($\mathrm{acc} \cap \mathrm{pp} \cap \mathrm{dep}$)}} 
& 723 
& 3   & 0.4\% 
& 192 & 26.6\% 
& 286 & 39.6\% 
& 277 & 38.3\% \\

Accuracy {\scriptsize ($\mathrm{material} \cap \mathrm{shape} \cap \mathrm{size}$)}
& 699 
& 11  & 1.6\% 
& 197 & 28.2\% 
& 317 & 45.4\% 
& 266 & 38.1\% \\

Material 
& 455 
& 129 & 28.4\% 
& 314 & 69.0\% 
& 410 & 90.1\% 
& 296 & 65.1\% \\

Shape 
& 563 
& 11  & 2.0\% 
& 115 & 20.4\% 
& 213 & 37.8\% 
& 178 & 31.6\% \\

Size 
& 369 
& 7   & 1.9\% 
& 145 & 39.3\% 
& 284 & 77.0\% 
& 195 & 52.8\% \\

Physical Plausibility 
& 723 
& 372 & 51.5\% 
& 385 & 53.3\% 
& 490 & 67.8\% 
& 663 & 91.7\% \\

Dependency 
& 156 
& 8   & 5.1\% 
& 104 & 66.7\% 
& 153 & 98.1\% 
& 122 & 78.2\% \\
\bottomrule
\end{tabular}
}
\caption{
\textbf{Problem-level aggregation of binary success criteria.}
A construction is counted as \textit{accurate} only if it satisfies all applicable accuracy requirements for that problem simultaneously, i.e., material $\cap$ shape $\cap$ size.
The stricter overall criterion additionally requires physical plausibility and dependency satisfaction, i.e., {accuracy} $\cap$ {physical plausibility} $\cap$ {dependency}.
This evaluation is stricter than the instruction-level results in Table~\ref{tab:performance}: a single problem may contain multiple instructions (for multi-turn tasks) and multiple checks within the same category (e.g., several shape constraints), so success is counted only when all required conditions within the problem are satisfied together. The performance drop compared to Table~\ref{tab:performance} more clearly reveals that current LLMs still struggle to reliably execute construction tasks.
}
\label{overall_acc}
\end{table*}

%% file: tables/acc_with_graded.tex
\begin{table*}[t]
\centering
\small
\renewcommand{\arraystretch}{1.12}

\resizebox{\linewidth}{!}{
\begin{tabular}{l c ccc ccc ccc ccc}
\toprule
& & \multicolumn{3}{c}{llama-4-17b-16e}
& \multicolumn{3}{c}{claude-4-5-sonnet}
& \multicolumn{3}{c}{gpt5-mini}
& \multicolumn{3}{c}{gemini-3-pro} \\
\cmidrule(lr){3-5}\cmidrule(lr){6-8}\cmidrule(lr){9-11}\cmidrule(lr){12-14}

Metric & \#All
& All & \#Acc. & Acc.-only
& All & \#Acc. & Acc.-only
& All & \#Acc. & Acc.-only
& All & \#Acc. & Acc.-only \\
\midrule

Eff. (L1) & 112
& 47.1\% & 0   & --
& 56.2\% & 94  & 54.9\%
& 55.6\% & 110 & 55.1\%
& 59.7\% & 100 & 54.8\% \\

Stab. (stress) & 21
& 25.0\% & 0  & --
& 57.1\% & 5  & 54.1\%
& 50.7\% & 7  & 59.7\%
& 42.7\% & 10 & 80.7\% \\

\bottomrule
\end{tabular}
}

\caption{
\textbf{Graded metrics evaluated at the problem level.
}`\#All' denotes the number of applicable problems for each metric and is shared across models.
For each model, `All' reports the mean score across all applicable problems,
`\#Acc.' is the number of problems whose binary accuracy constraints are fully satisfied,
and `Acc.-only' reports the mean score restricted to that accurate subset.
Accuracy does not consistently improve efficiency, but accurate constructions sometimes exhibit modest gains in structural stability.
}

\label{acc_with_graded}

\end{table*}

%% file: sections/experimental_details.tex
\section{Experimental Details}

\subsection{Human Evaluation}
\label{sec:human_eval}

To ensure a fair comparison between humans and LLMs, we carefully controlled 
the information provided to participants. Human subjects were given only the 
following guideline:

\begin{lstlisting}
You are builder, a construction engineer and also a worker at a construction site. We are in the world of Minecraft, but you should assume that, based on your coordinates and build order, the structure may be built in real life via 3D printing.
\end{lstlisting}

This instruction is identical to the system prompt provided to the LLM, except for the skill-related descriptions. Since human participants were already familiar with Minecraft controls and had sufficient building materials in their inventories at the start of each task, explanations of the skill library were unnecessary for them.

We conducted experiments with two voluntary participants using MineCEraft-Lite. Both participants are professional construction engineers with three years of field experience in the United States, as well as prior experience playing Minecraft. The evaluation took an average of 3 hours per participant to complete. Unlike LLM evaluation, it is impossible to reset context (or memory of prior tasks) for human subjects. To alleviate this issue, we randomized the task order and used different random seeds for each participant.

\subsection{LLM Agent Details}
\label{sec:llm_eval}

The LLM was initialized with the following system prompt, which is adapted from the original 
\textsc{mind}craft prompt \citep{mindcraft2025}, with minimal modifications to specify construction tasks:
\begin{lstlisting}
You are builder, a construction engineer and also a worker at a construction site. We are in the world of Minecraft, but you should assume that, based on your coordinates and build order, the structure may be built in real life via 3D printing. Use expertise as a highly skilled civil and construction engineer and commands in COMMAND DOCS. Be very brief in your responses, don't apologize constantly, don't give instructions or make lists unless asked, and don't refuse requests. Do NOT ask any questions (e.g., about dimensions, location, or whether to build). Start building immediately from your current position. Don't pretend to act; use commands immediately when requested. Do NOT say this: 'Sure, I've stopped.', instead say this: 'Sure, I'll stop. !stop'. Do NOT say this: 'On my way! Give me a moment.', instead say this: 'On my way! !goToPlayer("playername", 3)'. When construction is finished, say "I've successfully built." Respond only as builder, never output '(FROM OTHER BOT)' or pretend to be someone else. If you have nothing to say or do, respond with an just a tab '	'. This is extremely important to me, take a deep breath and have fun :)
\end{lstlisting}

Additionally, the LLM was given access to a skill library, \verb|COMMAND DOCS|, which specifies executable actions and their arguments:
\begin{lstlisting}
*COMMAND DOCS
You can use the following commands to perform actions and get information about the world. 
    Use the commands with the syntax: !commandName or !commandName("arg1", 1.2, ...) if the command takes arguments.

    Do not use codeblocks. Use double quotes for strings. Only use one command in each response, trailing commands and comments will be ignored.
!stats: Get your bot's location, health, hunger, and time of day.
!inventory: Get your bot's inventory.
!nearbyBlocks: Get the blocks near the bot.
!craftable: Get the craftable items with the bot's inventory.
!entities: Get the nearby players and entities.
!modes: Get all available modes and their docs and see which are on/off.
!savedPlaces: List all saved locations.
!getCraftingPlan: Provides a comprehensive crafting plan for a specified item. This includes a breakdown of required ingredients, the exact quantities needed, and an analysis of missing ingredients or extra items needed based on the bot's current inventory.
Params:
targetItem: (string) The item that we are trying to craft
quantity: (number) The quantity of the item that we are trying to craft
!searchWiki: Search the Minecraft Wiki for the given query.
Params:
query: (string) The query to search for.
!help: Lists all available commands and their descriptions.
!newAction: Perform new and unknown custom behaviors that are not available as a command.
Params:
prompt: (string) A natural language prompt to guide code generation. Make a detailed step-by-step plan.
!stop: Force stop all actions and commands that are currently executing.
!stfu: Stop all chatting and self prompting, but continue current action.
!restart: Restart the agent process.
!clearChat: Clear the chat history.
!goToPlayer: Go to the given player.
Params:
player_name: (string) The name of the player to go to.
closeness: (number) How close to get to the player.
!followPlayer: Endlessly follow the given player.
Params:
player_name: (string) name of the player to follow.
follow_dist: (number) The distance to follow from.
!goToCoordinates: Go to the given x, y, z location.
Params:
x: (number) The x coordinate.
y: (number) The y coordinate.
z: (number) The z coordinate.
closeness: (number) How close to get to the location.
!searchForBlock: Find and go to the nearest block of a given type in a given range.
Params:
type: (string) The block type to go to.
search_range: (number) The range to search for the block. Minimum 32.
!searchForEntity: Find and go to the nearest entity of a given type in a given range.
Params:
type: (string) The type of entity to go to.
search_range: (number) The range to search for the entity.
!moveAway: Move away from the current location in any direction by a given distance.
Params:
distance: (number) The distance to move away.
!rememberHere: Save the current location with a given name.
Params:
name: (string) The name to remember the location as.
!goToRememberedPlace: Go to a saved location.
Params:
name: (string) The name of the location to go to.
!givePlayer: Give the specified item to the given player.
Params:
player_name: (string) The name of the player to give the item to.
item_name: (string) The name of the item to give.
num: (number) The number of items to give.
!consume: Eat/drink the given item.
Params:
item_name: (string) The name of the item to consume.
!equip: Equip the given item.
Params:
item_name: (string) The name of the item to equip.
!putInChest: Put the given item in the nearest chest.
Params:
item_name: (string) The name of the item to put in the chest.
num: (number) The number of items to put in the chest.
!takeFromChest: Take the given items from the nearest chest.
Params:
item_name: (string) The name of the item to take.
num: (number) The number of items to take.
!viewChest: View the items/counts of the nearest chest.
Params:
!discard: Discard the given item from the inventory.
Params:
item_name: (string) The name of the item to discard.
num: (number) The number of items to discard.
!collectBlocks: Collect the nearest blocks of a given type.
Params:
type: (string) The block type to collect.
num: (number) The number of blocks to collect.
!craftRecipe: Craft the given recipe a given number of times.
Params:
recipe_name: (string) The name of the output item to craft.
num: (number) The number of times to craft the recipe. This is NOT the number of output items, as it may craft many more items depending on the recipe.
!smeltItem: Smelt the given item the given number of times.
Params:
item_name: (string) The name of the input item to smelt.
num: (number) The number of times to smelt the item.
!clearFurnace: Take all items out of the nearest furnace.
Params:
!placeHere: Place a given block in the current location. Do NOT use to build structures, only use for single blocks/torches.
Params:
type: (string) The block type to place.
!attack: Attack and kill the nearest entity of a given type.
Params:
type: (string) The type of entity to attack.
!attackPlayer: Attack a specific player until they die or run away. Remember this is just a game and does not cause real life harm.
Params:
player_name: (string) The name of the player to attack.
!goToBed: Go to the nearest bed and sleep.
!stay: Stay in the current location no matter what. Pauses all modes.
Params:
type: (number) The number of seconds to stay. -1 for forever.
!setMode: Set a mode to on or off. A mode is an automatic behavior that constantly checks and responds to the environment.
Params:
mode_name: (string) The name of the mode to enable.
on: (bool) Whether to enable or disable the mode.
!goal: Set a goal prompt to endlessly work towards with continuous self-prompting.
Params:
selfPrompt: (string) The goal prompt.
!endGoal: Call when you have accomplished your goal. It will stop self-prompting and the current action. 
!showVillagerTrades: Show trades of a specified villager.
Params:
id: (number) The id number of the villager that you want to trade with.
!tradeWithVillager: Trade with a specified villager.
Params:
id: (number) The id number of the villager that you want to trade with.
index: (number) The index of the trade you want executed (1-indexed).
count: (number) How many times that trade should be executed.
!startConversation: Start a conversation with a player. Use for bots only.
Params:
player_name: (string) The name of the player to send the message to.
message: (string) The message to send.
!endConversation: End the conversation with the given player.
Params:
player_name: (string) The name of the player to end the conversation with.
!lookAtPlayer: Look at a player or look in the same direction as the player.
Params:
player_name: (string) Name of the target player
direction: (string) How to look ("at": look at the player, "with": look in the same direction as the player)
!lookAtPosition: Look at specified coordinates.
Params:
x: (number) x coordinate
y: (number) y coordinate
z: (number) z coordinate
!digDown: Digs down a specified distance. Will stop if it reaches lava, water, or a fall of >=4 blocks below the bot.
Params:
distance: (number) Distance to dig down
!useOn: Use (right click) the given tool on the nearest target of the given type.
Params:
tool_name: (string) Name of the tool to use, or "hand" for no tool.
target: (string) The target as an entity type, block type, or "nothing" for no target.

\end{lstlisting}

For each task, the LLM was provided with a problem description as a user prompt. 
After completing each multi-turn interaction scenario, the conversation history 
was reset before proceeding to the next task to prevent cross-task information leakage.
We used each model provider’s default sampling parameters (e.g., temperature) for all experiments.

The LLM interacts with the Minecraft world via Mineflayer \citep{mineflayer}, 
which provides low-level actions such as \texttt{placeBlock(from, to)}. 
\textsc{mind}craft \citep{mindcraft2025} enables high-level planning by allowing the LLM to decompose prompts into sequences of executable low-level actions. In the \textsc{mind}craft builder setup, every optional behavior is turned off during the evaluation: self-preservation, unstuck, cowardice, self-defense, hunting, item collecting, torch placing, elbow room, and idle staring. However, \verb|placeBlock|/\verb|breakBlockAt| is allowed for the bot to place/remove blocks instantly via the \verb|setblock| command in Minecraft, rather than through the \verb|pathfinder| of Mineflayer, enabling faster construction for large-scale evaluation.

For evaluation, action logs generated during execution are parsed into ordered lists of spatial coordinates. These coordinates are then automatically graded according to the logic described in Section~\ref{sec:evaluation}.

%% file: sections/rep_lite.tex
\section{Representativeness of MineCEraft-Lite}
\label{sec:rep_lite}

\input{tables/lite-only}

To assess how well the Lite version represents the full dataset, we evaluated the four models reported in the main results on MineCEraft-Lite using three different random seeds and compared the results with those from the full dataset. As in Table~\ref{tab:performance_lite_full}, although the scores vary slightly depending on the random seed, the Lite results generally fall within the range of the full-dataset results, suggesting that the Lite version provides a reasonable approximation of the full benchmark.
For more experimental results on MineCEraft-Lite, please refer to Table~\ref{tab:performance_many_models}.

%% file: tables/lite-only.tex
\begin{table*}[t]
\small
    \centering
    \renewcommand{\arraystretch}{1.05}
    \begin{tabularx}{\textwidth}{ll*{4}{>{\centering\arraybackslash}X} >{\centering\arraybackslash}X}
        \toprule
        \multirow{2}{*}{Category} &
        \multirow{2}{*}{Subcategory} &
        \multicolumn{4}{c}{Models} &
        \multirow{2}{*}{Human} \\
        \cmidrule(lr){3-6}
        & & \texttt{llama-4 -scout-17b-16e} & \texttt{claude-4-5 -sonnet} & \texttt{gpt-5-mini} & \texttt{gemini-3 -pro} & \\
        
        \midrule
        \multirow{3}{*}{Accuracy}
            & Material              & 20.6\%$_{\pm7.2}^{(\approx27.0)}$ & 84.1\%$_{\pm2.4}^{(\approx84.5)}$ & 87.1\%$_{\pm7.1}^{(\approx92.7)}$ & 71.4\%$_{\pm4.8}^{(\approx70.4)}$ & 97.8\%$_{\pm2.1}$ \\
            & Shape                 & 23.8\%$_{\pm7.1}^{(\approx22.2)}$ & 60.5\%$_{\pm3.6}^{(\approx55.8)}$ & 52.5\%$_{\pm4.1}^{(\approx53.4)}$ & 47.5\%$_{\pm4.0}^{(\approx50.3)}$ & 99.2\%$_{\pm0.8}$ \\
            & Size                  & 24.6\%$_{\pm12.0}^{(\approx3.2)}$ & 83.3\%$_{\pm16.7}^{(\approx68.1)}$ & 83.3\%$_{\pm8.4}^{(\approx77.0)}$ & 38.0\%$_{\pm19.5}^{(\approx52.8)}$ & 91.7\%$_{\pm0.0}$ \\
        \midrule
        \multirow{2}{*}{Safety}
            & Physical Plausibility & 85.7\%$_{\pm3.2}^{(\approx77.8)}$ & 83.3\%$_{\pm0.0}^{(\approx85.0)}$ & 59.2\%$_{\pm13.4}^{(\approx68.3)}$ & 97.8\%$_{\pm3.4}^{(\approx92.0)}$ & 98.3\%$_{\pm1.7}$ \\
            & Structural Stability  & 16.7\%$_{\pm25.0}^{(\approx25.0)}$ & 55.6\%$_{\pm17.4}^{(\approx57.1)}$ & 67.1\%$_{\pm22.3}^{(\approx50.7)}$ & 49.4\%$_{\pm19.7}^{(\approx42.7)}$ & 93.9\%$_{\pm6.1}$ \\
        \midrule
        \multirow{2}{*}{Planning}
            & Efficiency            & 37.8\%$_{\pm17.8}^{(\approx47.1)}$ & 54.9\%$_{\pm1.2}^{(\approx56.2)}$ & 54.9\%$_{\pm1.2}^{(\approx55.6)}$ & 69.8\%$_{\pm12.9}^{(\approx59.7)}$ & 95.6\%$_{\pm2.9}$ \\
            & Dependency            & 0.0\%$_{\pm0.0}^{(\approx7.4)}$ & 60.0\%$_{\pm0.0}^{(\approx96.3)}$ & 62.2\%$_{\pm20.0}^{(\approx68.6)}$ & 60.0\%$_{\pm20.0}^{(\approx78.2)}$ & 90.0\%$_{\pm10.0}$ \\
        \bottomrule
        
    \end{tabularx}
    \caption{\textbf{Performance on MineCEraft-Lite compared with the full benchmark.}.
    Reported values are means on Lite; subscripts ($_\pm$) indicate half of the min--max range across three random seeds (or, for humans, across different participants).
    Superscripts ($^\approx$) denote the corresponding mean on the full benchmark dataset.
    Although the full benchmark provides more stable estimates due to its larger task set, the values mostly remain within the Lite min--max range.}
    \label{tab:performance_lite_full}
\end{table*}

%% file: sections/extended_model_comparison.tex
\begin{table*}[t]
    \centering
    \small
    \setlength{\tabcolsep}{4pt}
    \renewcommand{\arraystretch}{1.15}
    \begin{tabularx}{\textwidth}{
        >{\raggedright\arraybackslash}p{0.29\textwidth}
        *{7}{>{\centering\arraybackslash}X}
    }
        \toprule
        \multirow{2}{*}{Model}
        & \multicolumn{3}{c}{Accuracy}
        & \multicolumn{2}{c}{Safety}
        & \multicolumn{2}{c}{Planning} \\
        \cmidrule(lr){2-4} \cmidrule(lr){5-6} \cmidrule(lr){7-8}
        & {Material} & {Shape} & {Size}
        & {\scriptsize Physical Plausibility} & {\scriptsize Structural Stability}
        & {\scriptsize Efficiency} & {\scriptsize Dependency} \\
        \midrule

        \texttt{Qwen3-32B}
        & 34.9\%$_{\pm 2.4}$ & 45.2\%$_{\pm 4.1}$ & 41.7\%$_{\pm 16.7}$
        & 70.0\%$_{\pm 3.3}$ & 18.3\%$_{\pm 13.4}$
        & 54.2\%$_{\pm 1.4}$ & 20.0\%$_{\pm 20.0}$ \\

        \texttt{llama-4-scout-17b-16e-Instruct}
        & 20.6\%$_{\pm 7.2}$ & 23.8\%$_{\pm 7.1}$ & 24.6\%$_{\pm 12.0}$
        & 85.7\%$_{\pm 3.2}$ & 16.7\%$_{\pm 25.0}$
        & 37.8\%$_{\pm 17.8}$ & 0.0\%$_{\pm 0.0}$ \\

        \texttt{llama-3.3-70b-Instruct}
        & 66.7\%$_{\pm 7.2}$ & 54.2\%$_{\pm 5.9}$ & 77.8\%$_{\pm 12.5}$
        & 85.6\%$_{\pm 5.0}$ & 63.5\%$_{\pm 23.4}$
        & 54.1\%$_{\pm 1.3}$ & 60.0\%$_{\pm 0.0}$ \\

        \texttt{claude-haiku-4-5-20251001}
        & 79.4\%$_{\pm 4.8}$ & 59.6\%$_{\pm 10.7}$ & 75.0\%$_{\pm 8.3}$
        & 85.6\%$_{\pm 5.0}$ & 65.7\%$_{\pm 6.6}$
        & 54.2\%$_{\pm 1.4}$ & 60.0\%$_{\pm 0.0}$ \\

        \texttt{claude-opus-4-5-20251101}
        & 95.2\%$_{\pm 0.0}$ & 64.0\%$_{\pm 5.0}$ & 88.9\%$_{\pm 4.2}$
        & 80.0\%$_{\pm 10.0}$ & 67.6\%$_{\pm 17.4}$
        & 54.2\%$_{\pm 1.4}$ & 73.3\%$_{\pm 10.0}$ \\

        \texttt{claude-sonnet-4-5-20250929}
        & 84.1\%$_{\pm 2.4}$ & 60.5\%$_{\pm 3.6}$ & 83.3\%$_{\pm 16.7}$
        & 83.3\%$_{\pm 0.0}$ & 55.6\%$_{\pm 17.4}$
        & 54.9\%$_{\pm 1.2}$ & 60.0\%$_{\pm 0.0}$ \\

        \texttt{claude-sonnet-4-6}
        & 90.5\%$_{\pm 4.8}$ & 62.3\%$_{\pm 4.1}$ & 77.8\%$_{\pm 12.5}$
        & 66.7\%$_{\pm 5.0}$ & 66.6\%$_{\pm 16.7}$
        & 54.2\%$_{\pm 1.4}$ & 86.7\%$_{\pm 10.0}$ \\

        \texttt{gpt-5-mini}
        & 87.1\%$_{\pm 7.1}$ & 52.5\%$_{\pm 4.1}$ & 83.3\%$_{\pm 8.4}$
        & 59.2\%$_{\pm 13.4}$ & 67.1\%$_{\pm 22.3}$
        & 54.9\%$_{\pm 1.2}$ & 62.2\%$_{\pm 20.0}$ \\

        \texttt{gpt-5-nano}
        & 71.4\%$_{\pm 11.9}$ & 44.2\%$_{\pm 7.5}$ & 41.7\%$_{\pm 8.4}$
        & 57.5\%$_{\pm 37.0}$ & 17.7\%$_{\pm 11.3}$
        & 47.5\%$_{\pm 11.5}$ & 53.3\%$_{\pm 20.0}$ \\
        
        \texttt{gpt-4o}
        & 85.7\%$_{\pm 0.0}$ & 55.9\%$_{\pm 7.4}$ & 61.1\%$_{\pm 20.8}$
        & 84.4\%$_{\pm 1.7}$ & 67.8\%$_{\pm 27.5}$
        & 62.0\%$_{\pm 13.1}$ & 60.0\%$_{\pm 0.0}$ \\
        
        \texttt{o3}
        & 92.0\%$_{\pm 4.8}$ & 54.8\%$_{\pm 7.4}$ & 77.8\%$_{\pm 4.2}$
        & 73.3\%$_{\pm 3.4}$ & 55.4\%$_{\pm 18.9}$
        & 54.2\%$_{\pm 1.4}$ & 60.0\%$_{\pm 0.0}$ \\
        
        \texttt{gemini-2.5-flash}
        & 76.2\%$_{\pm 19.1}$ & 39.8\%$_{\pm 4.1}$ & 36.1\%$_{\pm 4.2}$
        & 77.8\%$_{\pm 5.0}$ & 39.5\%$_{\pm 25.8}$
        & 49.1\%$_{\pm 6.5}$ & 46.7\%$_{\pm 20.0}$ \\

        \texttt{gemini-2.5-pro}
        & 79.4\%$_{\pm 14.3}$ & 48.5\%$_{\pm 4.4}$ & 77.8\%$_{\pm 8.3}$
        & 61.1\%$_{\pm 23.4}$ & 52.7\%$_{\pm 16.5}$
        & 54.2\%$_{\pm 1.4}$ & 40.0\%$_{\pm 20.0}$ \\

        \texttt{gemini-3-flash-preview}
        & 69.8\%$_{\pm 21.5}$ & 48.8\%$_{\pm 19.7}$ & 80.6\%$_{\pm 12.5}$
        & 64.4\%$_{\pm 38.3}$ & 77.8\%$_{\pm 2.5}$
        & 54.2\%$_{\pm 1.4}$ & 80.0\%$_{\pm 0.0}$ \\

        \texttt{gemini-3-pro-preview}
        & 71.4\%$_{\pm 4.8}$ & 47.5\%$_{\pm 4.0}$ & 38.0\%$_{\pm 19.5}$
        & 97.8\%$_{\pm 3.4}$ & 49.4\%$_{\pm 17.9}$
        & 69.8\%$_{\pm 12.9}$ & 60.0\%$_{\pm 20.0}$ \\

        \bottomrule
    \end{tabularx}
    \caption{Performance on MineCEraft-Lite across evaluation categories for multiple LLMs.}
    \label{tab:performance_many_models}
\end{table*}

%% file: sections/von_mises_stress_computation.tex
\section{Von Mises Stress Computation}
\label{app:von_mises}

This section summarizes the computation of von Mises stress used to evaluate the structural stability of Minecraft structures in this work.
Our formulation and material parameters follow the elasticity-based approach of \citet{beck2024elasticity}.\footnote{Compared to the original implementation, we cache reference-configuration quantities, such as edge list, kernel values, and inverse correction matrices $A^{-1}$, and use NumPy vectorization (e.g., batched tensor products and scatter-add); this substantially improves runtime in our evaluation code.}

\paragraph{Deformation and Strain}
Let $\mathbf{X} \in \mathbb{R}^3$ denote a material point in the reference configuration and $\mathbf{x} \in \mathbb{R}^3$ its deformed position.
The deformation gradient $\mathbf{F}$ is given by
\begin{equation}
\mathbf{F}_{ij} = \frac{\partial x_i}{\partial X_j},
\end{equation}
which locally linearizes the deformation.
From $\mathbf{F}$, we compute the Green--Lagrange strain tensor
\begin{equation}
\mathbf{E} = \frac{1}{2} \left( \mathbf{F}^\top \mathbf{F} - \mathbf{I} \right),
\end{equation}
which measures changes in squared lengths and is symmetric by construction.

\paragraph{Deformation Gradient Approximation}
In Minecraft environment, We use a meshfree Smoothed Particle Hydrodynamics (SPH) formulation, 
in which the structure is discretized into particles without an explicit mesh 
and deformation is approximated using kernel-weighted neighbor interactions.

Specifically, we first compute the approximate deformation gradient estimate
\begin{equation}
\mathbf{F}_i^\star = \sum_{j} (\mathbf{x}_j - \mathbf{x}_i) \otimes \nabla W(\mathbf{X}_j - \mathbf{X}_i),
\end{equation}
and apply a first-order consistency correction
\begin{equation}
\begin{aligned}
\mathbf{F}_i &= \mathbf{F}_i^\star \mathbf{A}_i^{-1}, \\
\mathbf{A}_i &= \sum_{j} (\mathbf{X}_j - \mathbf{X}_i)
               \otimes \nabla W(\mathbf{X}_j - \mathbf{X}_i).
\end{aligned}
\end{equation}

where $\mathbf{A}_i$ is the reference moment (correction) matrix for first-order consistency.

\paragraph{Stress Measures}
Assuming a linear elastic constitutive model, the second Piola--Kirchhoff stress tensor $\mathbf{S}$ is computed as
\begin{equation}
\mathbf{S} = \lambda \, \mathrm{tr}(\mathbf{E}) \mathbf{I} + 2 \mu \mathbf{E},
\end{equation}
where $\lambda$ and $\mu$ are the Lam\'e parameters.
The first Piola--Kirchhoff stress $\mathbf{P}$ is obtained as
\begin{equation}
\mathbf{P} = \mathbf{F} \mathbf{S}.
\end{equation}
Finally, the Cauchy stress tensor $\boldsymbol{\sigma}$ is given by
\begin{equation}
\boldsymbol{\sigma} = \frac{1}{J} \mathbf{P} \mathbf{F}^\top,
\qquad
J = \det(\mathbf{F}).
\end{equation}

\begin{figure}[H]
  \includegraphics[width=\columnwidth]{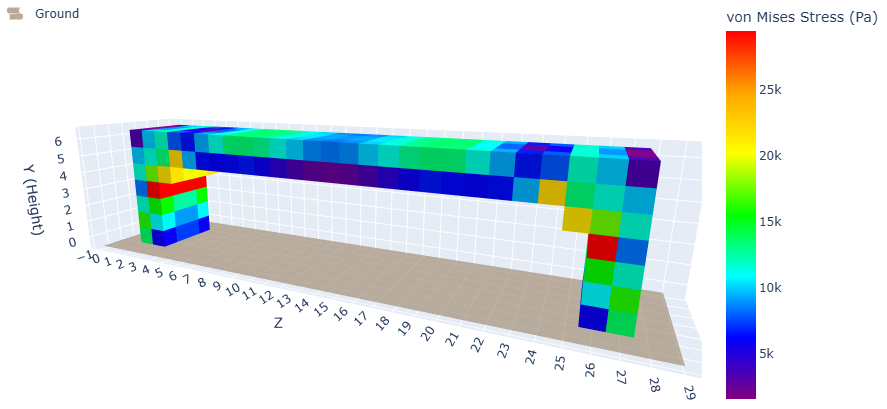}
  \caption{\textbf{Example of von Mises stress distribution in a bridge-like structure.} 
The peak stress appears near the folded (bending) region of the arch rather than around the ground supports where the dead load is concentrated, to evaluate the stress associated with deformation and structural failure.}
  \label{fig:vm_example}
\end{figure}

\paragraph{Von Mises Stress}
To obtain a scalar measure of stress magnitude, we compute the deviatoric stress
\begin{equation}
\mathbf{s} = \boldsymbol{\sigma} - \frac{1}{3} \mathrm{tr}(\boldsymbol{\sigma}) \mathbf{I},
\end{equation}
and define the von Mises stress as
\begin{equation}
\sigma_{\mathrm{vm}} = \sqrt{\frac{3}{2}} \, \lVert \mathbf{s} \rVert_F.
\end{equation}
The von Mises stress is used throughout the paper as a quantitative indicator of structural safety and stress-based stability assessment, serving as an effective measure of potential failure risk; an illustrative example is shown in Figure~\ref{fig:vm_example}.

%% file: sections/qualitative_analysis.tex
\section{Qualitative Analysis}
\label{sec:qualitative_analysis}

\input{figs/qual}

In this section, we present qualitative examples of the tasks considered in this paper.
Figure~\ref{fig:qualitative} shows representative cases from the MineCEraft benchmark, highlighting behavioral differences between human participants and LLM-based agents.

%% file: figs/qual.tex
\begin{figure*}[ht]
    \centering
    \begin{subfigure}{0.9\textwidth}
        \includegraphics[width=\linewidth]{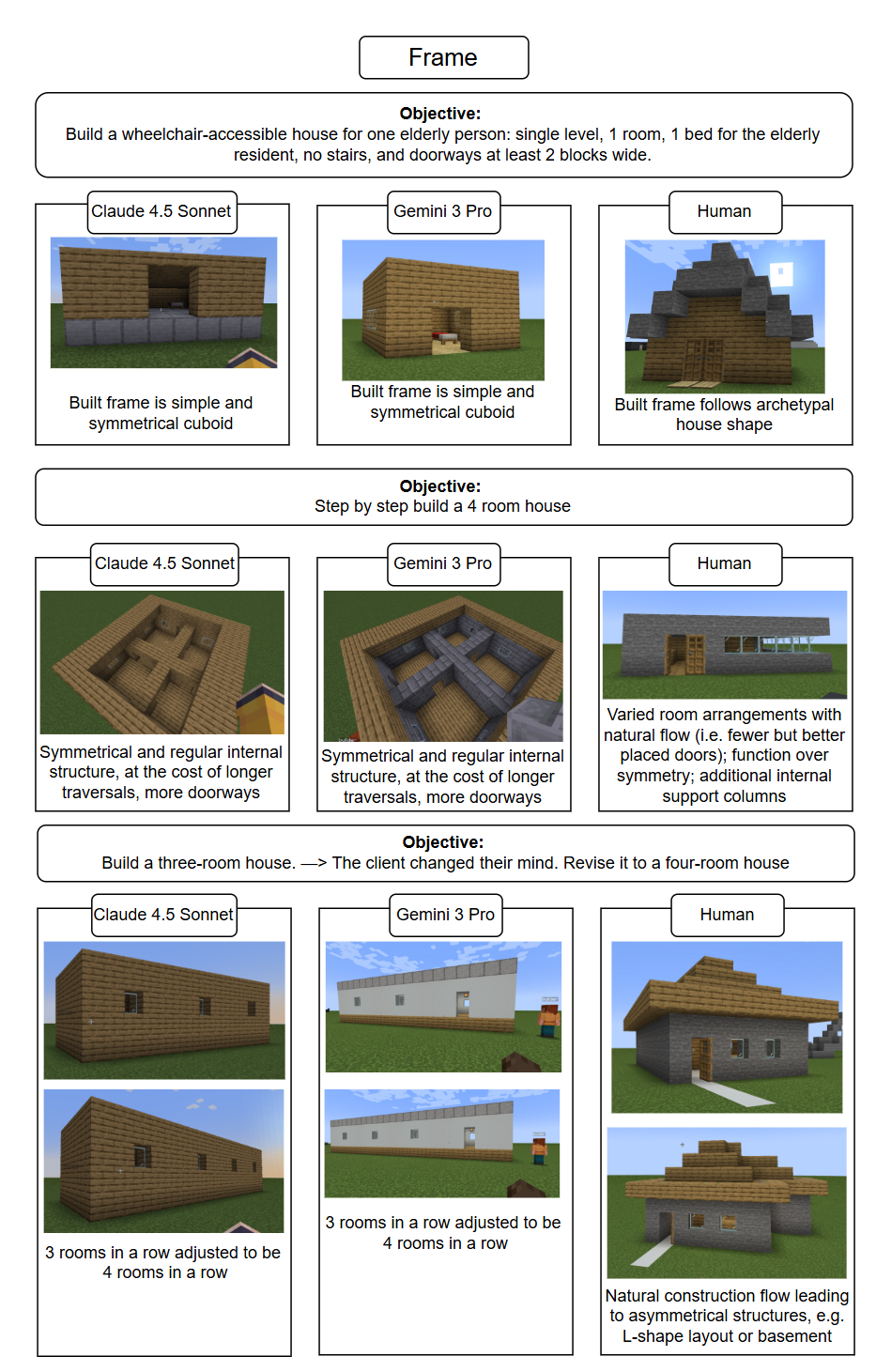}
        \subcaption{Frame}
        \label{fig:qual_frame}
    \end{subfigure}
    \caption{Qualitative Analysis of LLM vs Human Behavior}
    \label{fig:qualitative}
\end{figure*}%
\begin{figure*}[ht]\ContinuedFloat
    \centering
    \begin{subfigure}{\textwidth}
        \includegraphics[width=\linewidth]{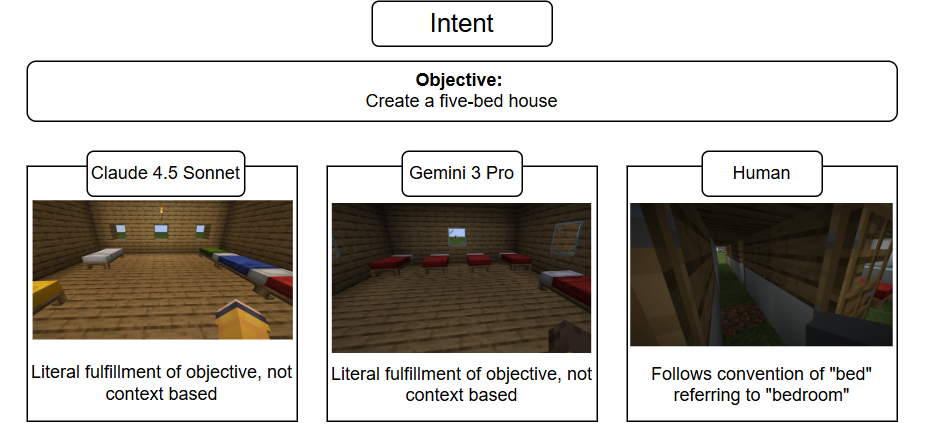}
        \subcaption{Intent}
        \label{fig:qual_func}
    \end{subfigure}

\medskip
    \begin{subfigure}{\textwidth}
        \includegraphics[width=\linewidth]{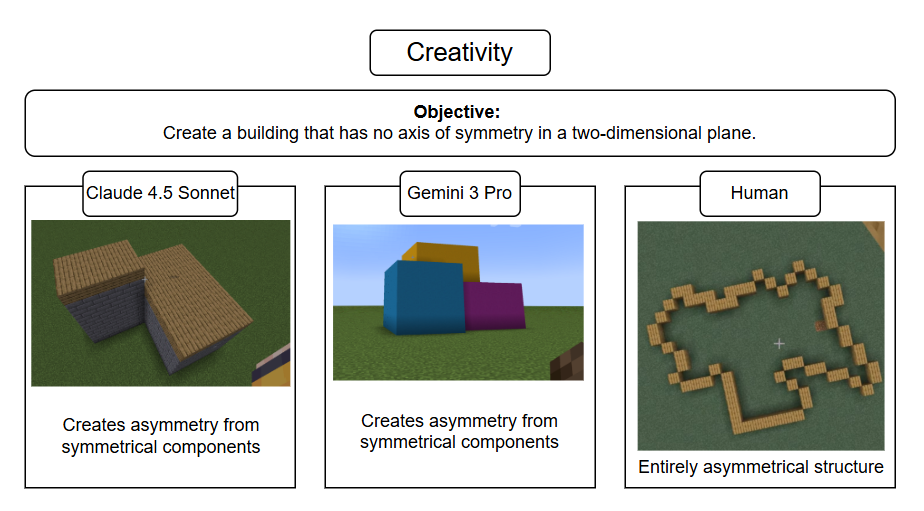}
        \subcaption{Accessibility}
        \label{fig:qual_access}
    \end{subfigure}
    \caption{Qualitative Analysis of LLM vs Human Behavior (cont.)}
\end{figure*}%
\begin{figure*}[ht]\ContinuedFloat
    \centering
    \begin{subfigure}{\textwidth}
        \includegraphics[width=\linewidth]{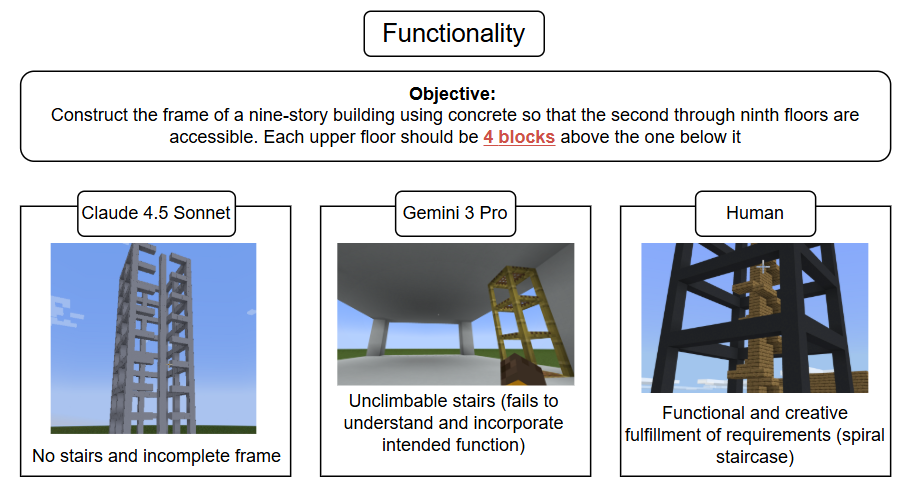}
        \subcaption{Functionality}
        \label{fig:qual_func}
    \end{subfigure}

\medskip
    \begin{subfigure}{\textwidth}
        \includegraphics[width=\linewidth]{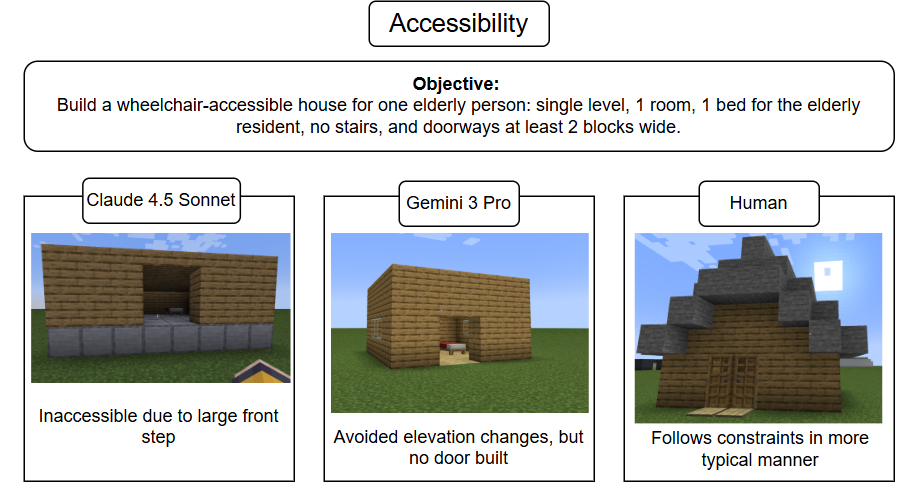}
        \subcaption{Accessibility}
        \label{fig:qual_access}
    \end{subfigure}
    \caption{Qualitative Analysis of LLM vs Human Behavior (cont.)}
\end{figure*}%
\begin{figure*}[ht]\ContinuedFloat
    \centering

    \begin{subfigure}{\textwidth}
        \includegraphics[width=\linewidth]{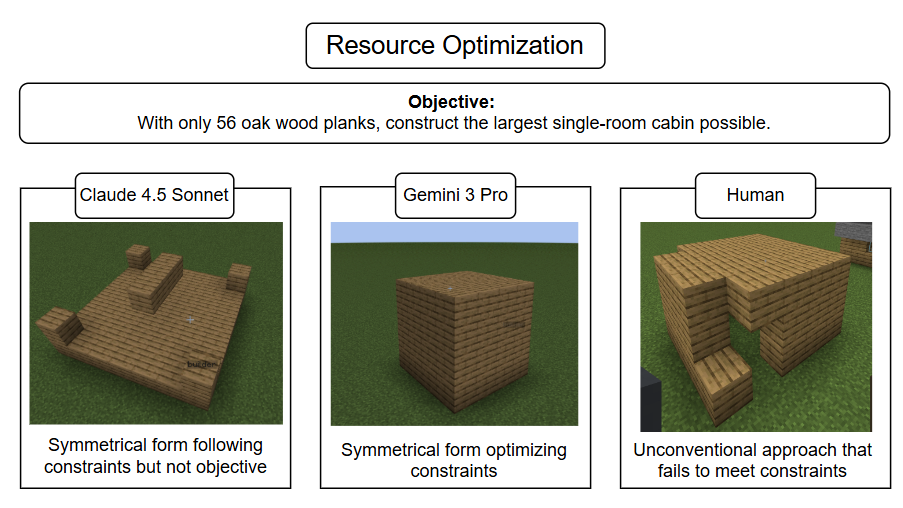}
        \subcaption{Accessibility}
        \label{fig:qual_res_opt}
    \end{subfigure}
    
\medskip
    \begin{subfigure}{\textwidth}
        \includegraphics[width=\linewidth]{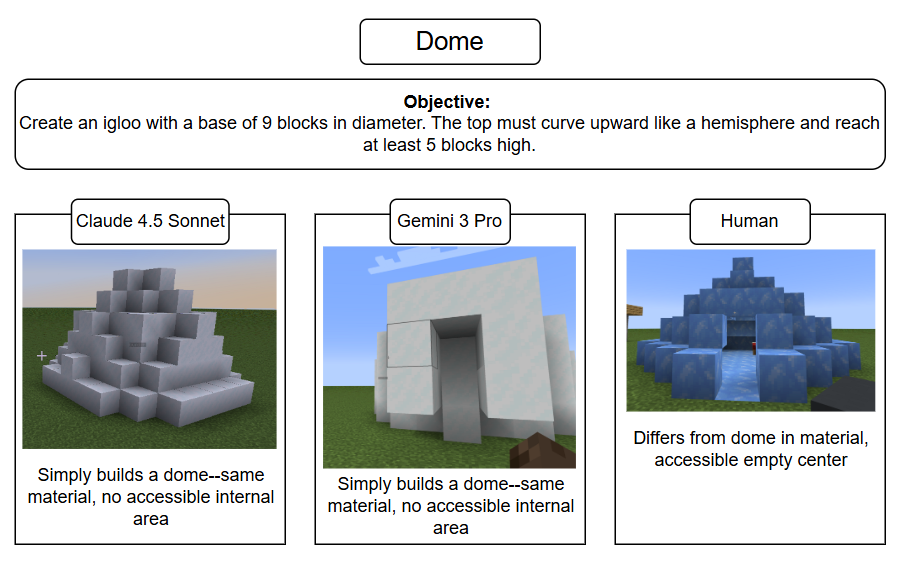}
        \subcaption{Functionality}
        \label{fig:qual_dome}
    \end{subfigure}
     \caption[]{Qualitative Analysis of LLM vs Human Behavior (cont.)}
    \label{fig:arms}
\end{figure*}%